\documentclass[final]{cvpr}

\usepackage[title]{appendix}
\usepackage{amsmath}
\usepackage{amssymb}
\usepackage{booktabs}
\usepackage{color}
\usepackage{colortbl}
\usepackage{comment}
\usepackage{enumitem}
\usepackage{epsfig}
\usepackage{etoolbox}
\usepackage{graphicx}
\usepackage[utf8]{inputenc}
\usepackage{listings}
\usepackage{multirow}
\usepackage[numbers,sort&compress]{natbib}
\usepackage[olditem,oldenum]{paralist}
\usepackage{subfig}
\usepackage{tabularx}
\usepackage{times}
\usepackage{transparent}
\usepackage[dvipsnames]{xcolor}

\usepackage[pagebackref,breaklinks,colorlinks,bookmarks=False]{hyperref}
\usepackage{cleveref}

\usepackage{inconsolata}

\hypersetup{
   urlcolor=blue,
   citecolor=ForestGreen,
}

\def\cvprPaperID{410}
\def\confYear{CVPR 2021}

\newcommand{\virtex}[0]{VirTex}

\newcommand{\imagenet}[0]{ImageNet-1k}
\newcommand{\voc}[0]{PASCAL VOC}
\newcommand{\inclf}[0]{IN-1k}
\newcommand{\vocclf}[0]{VOC07}
\newcommand{\inat}[0]{iNaturalist 2018}
\newcommand{\inatclf}[0]{iNat 18}

\newcommand{\random}[0]{Random Init}
\newcommand{\insup}[0]{IN-sup}
\newcommand{\insupfif}[0]{IN-sup-$50\%$}
\newcommand{\insupten}[0]{IN-sup-$10\%$}
\newcommand{\mocoin}[0]{MoCo-IN}
\newcommand{\mocococo}[0]{MoCo-COCO}

\newcommand{\ttbf}[1]{\textbf{\texttt{#1}}}
\newcommand{\band}{\rowcolor{gray!20}}

\newcommand{\drop}[1]{\textcolor{gray}{\textsubscript{--#1}}}
\newcommand{\rise}[1]{\textcolor{gray}{\textsubscript{+#1}}}

\newcommand{\Drop}[1]{\textcolor{Red}{\textsubscript{\bf --#1}}}
\newcommand{\Rise}[1]{\textcolor{Green}{\textsubscript{\bf +#1}}}

\newcommand{\attention}[1]{\ttbf{\textcolor{ForestGreen}{#1}}}

\newcommand\YAMLcolonstyle{\color{Black}\mdseries}
\newcommand\YAMLkeystyle{\color{Black}\bfseries}
\newcommand\YAMLvaluestyle{\color{RoyalBlue}\mdseries}

\makeatletter
\newcommand\language@yaml{yaml}

\expandafter\lstdefinelanguage
\expandafter{\language@yaml}
{
  keywords={true,false,null},
  keywordstyle=\color{RoyalBlue},
  basicstyle=\YAMLkeystyle,
  sensitive=false,
  comment=[l]{\#},
  commentstyle=\color{RedOrange}\ttfamily,
  stringstyle=\YAMLvaluestyle\ttfamily,
  moredelim=**[il][\YAMLcolonstyle{:}\YAMLvaluestyle]{:},
  frame=tb,
}

\begin{document}

\title{VirTex: Learning Visual Representations from Textual Annotations}

\author{Karan Desai \qquad \qquad Justin Johnson\\
University of Michigan \\
{\tt\small \{kdexd,justincj\}@umich.edu}
}
\maketitle

\begin{abstract}
The de-facto approach to many vision tasks is to start from pretrained visual representations, typically learned via supervised training on ImageNet.
Recent methods have explored unsupervised pretraining to scale to vast quantities of unlabeled images.
In contrast, we aim to learn high-quality visual representations from fewer images.
To this end we revisit supervised pretraining, and seek data-efficient alternatives to classification-based pretraining.
We propose \virtex{} -- a pretraining approach using semantically dense captions to learn visual representations.
We train convolutional networks from scratch on COCO Captions, and transfer them to downstream recognition tasks including image classification, object detection, and instance segmentation.
On all tasks, \virtex{} yields features that match or exceed those learned on ImageNet -- supervised or unsupervised -- despite using up to ten times fewer images.

\end{abstract}

\section{Introduction}

The prevailing paradigm for learning visual representations is first to \emph{pretrain} a convolutional network~\cite{krizhevsky2012imagenet,he2016deep} to perform image classification on ImageNet~\cite{deng2009imagenet,russakovsky2015imagenet}, then \emph{transfer} the learned features to downstream tasks~\cite{donahue2014decaf,sharif2014cnn}.
This approach has been wildly successful, and has led to significant advances on a wide variety of computer vision problems such as object detection~\cite{girshick2014rich}, semantic~\cite{long2015fully} and instance~\cite{he2017mask} segmentation, image captioning~\cite{vinyals2015show,karpathy2015deep,donahue2015long}, and visual question answering~\cite{antol2015vqa,zhu2016visual7w}.
Despite its practical success, this approach is expensive to scale since the pretraining step relies on images annotated by human workers.

\begin{figure}[t!]
    \centering
    \includegraphics[width=\linewidth]{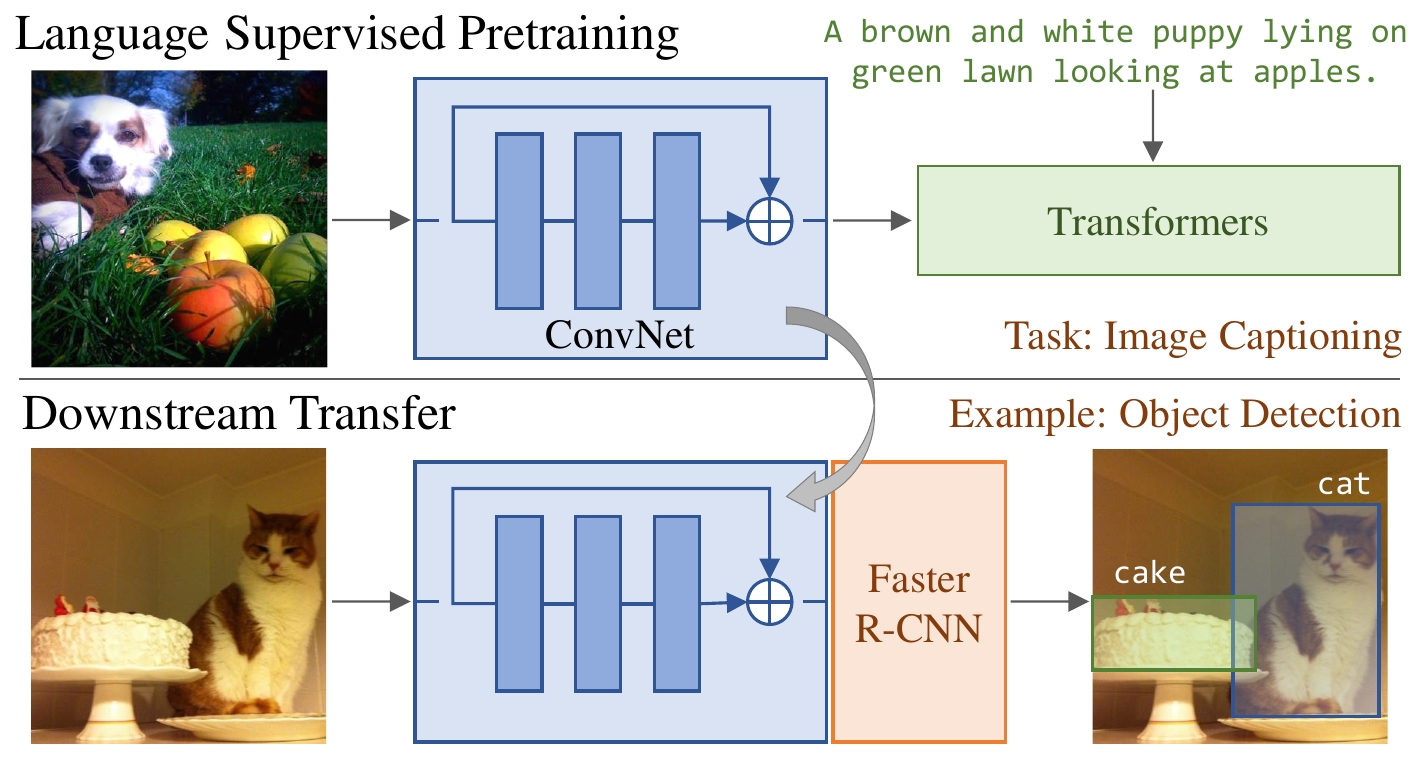}
    \vspace{-20pt}
    \caption{
        \textbf{Learning visual features from language:}
        First, we jointly train a ConvNet and Transformers using image-caption pairs, for the task of image captioning (top).
        Then, we transfer the learned ConvNet to several downstream vision tasks, for example object detection (bottom).
    }
    \vspace{-10pt}
    \label{fig:teaser}
\end{figure}

For this reason, there has been increasing interest in \emph{unsupervised pretraining} methods that use unlabeled images to learn visual representations which are then transferred to downstream tasks~\cite{doersch2015context,pathak2016inpainting,zhang2016colorful,zhang2017split,gidaris2018rotnet,oord2018cpcv1,tian2019cmc}.
Some recent approaches have begun to match or exceed supervised pretraining on ImageNet~\cite{goyal2019scaling,henaff2019cpcv2,he2019moco,misra2019pirl,chen2020simclr}, and have been scaled to hundreds of millions~\cite{caron2018deepcluster,goyal2019scaling,caron2019unsupervised,misra2019pirl} or billions~\cite{he2019moco} of images.

Continuing to scale unsupervised pretraining to ever-larger sets of unlabeled images is an important scientific goal.
But we may also ask whether there are alternate ways of pretraining that learn high-quality visual representations with \emph{fewer} images.
To do so, we revisit \emph{supervised} pretraining and seek an alternative to traditional classification pretraining that uses each image more efficiently.

In this paper we present an approach for learning \textbf{Vi}sual \textbf{r}epresentations from \textbf{Tex}tual annotations (\virtex{}).
Our approach is straightforward: first, we jointly train a ConvNet and Transformer~\cite{vaswani2017attention} \emph{from scratch} to generate natural language captions for images. Then, we transfer the learned features to downstream visual recognition tasks (\Cref{fig:teaser}).

\begin{figure*}[t!]
    \centering
    \includegraphics[width=0.96\textwidth]{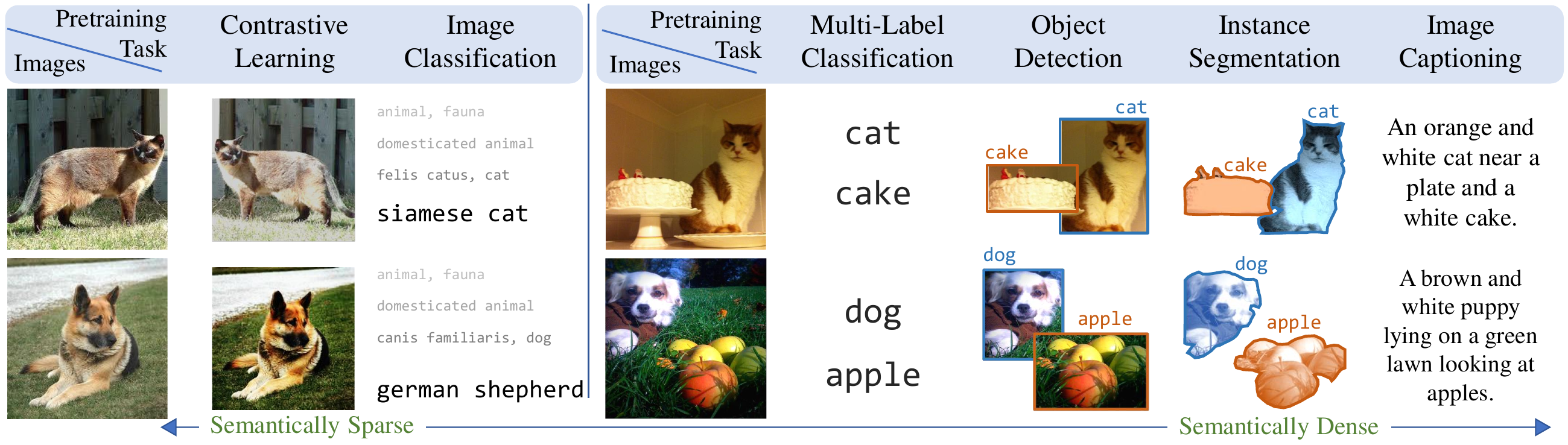}
    \vspace{-5pt}
    \caption{
        \textbf{Comparison of pretraining tasks for learning visual representations:}
        Contrastive self-supervised learning methods provide a \emph{semantically sparse} learning signal, encouraging different transforms of an image to have similar features.
        Image classification pairs an image with a single semantic concept, providing moderate semantic density.
        Multi-label classification, object detection, and instance segmentation increase semantic density by labeling and localizing multiple objects.
        Captions describe multiple objects, their attributes, relationships, and actions, giving a semantically dense learning signal.
        In this work, we aim to leverage this semantic density of captions to learn visual representations in a data-efficient manner.
    }
    \vspace{-10pt}
    \label{fig:comparison}
\end{figure*}

We believe that using language supervision is appealing due to its \emph{semantic density}.
Figure~\ref{fig:comparison} compares different pretraining tasks for learning visual representations.
Captions provide a semantically denser learning signal than unsupervised contrastive methods and supervised classification.
Hence, we expect that using textual features to learn visual features may require fewer images than other approaches.

Another benefit of textual annotations is simplified data collection.
To collect classification labels, typically human experts first build an ontology of categories~\cite{deng2009imagenet,russakovsky2015imagenet,lin2014microsoft,gupta2019lvis}, then complex crowdsourcing pipelines are used to elicit labels from non-expert users~\cite{deng2014scalable,krishna2016embracing}.
In contrast, natural language descriptions do not require an explicit ontology and can easily be written by non-expert workers, leading to a simplified data collection pipeline~\cite{hodosh2013framing,young2014image,chen2015microsoft}.
Large quantities of weakly aligned images and text can also be obtained from internet images~\cite{sharma2018conceptual,mao2016pinterest,ordonez2011im2text}.

Our main contribution is to show that natural language can provide supervision for learning transferable visual representations with better data-efficiency than other approaches. We train models from scratch on the COCO Captions dataset~\cite{chen2015microsoft}, and evaluate the learned features on downstream tasks including image classification, object detection, instance segmentation, and low-shot recognition.
On all tasks, \virtex{} matches or exceeds the performance of existing methods for supervised or unsupervised pretraining on ImageNet, despite using up to $10\times$ fewer images.
Our code and pretrained models are available at \texttt{\hyperlink{https://github.com/kdexd/virtex}{https://github.com/kdexd/virtex}}

\section{Related Work}
\label{sec:related}

Our work is related to recent efforts to move beyond supervised pretraining on ImageNet using alternate data sources or pretraining tasks.

\noindent \textbf{Weakly Supervised Learning}
scales beyond supervised pretraining with a \emph{quantity over quality} approach, and learns on large numbers of images with noisy labels from web services.
\emph{Li et al.}~\cite{li2017ngrams} trains visual N-gram models on the YFCC-100M dataset~\cite{yfcc100m}, that provides 100M Flickr images with user-provided tags.
Recent works~\cite{sun2017revisiting,kolesnikov2019large,xie2019self} also use JFT-300M~\cite{sun2017revisiting} dataset, curated by automatic labeling of images from web signals using Google's internal tooling.
Weakly-supervised learning has also been studied on up to 3.5B Instagram images, using hashtags as labels~\cite{mahajan2018exploring,yalniz2019billion}.
These approaches learn visual representations with large quantities of images with low-quality labels; in contrast we focus on using fewer images with high-quality annotations.

\noindent \textbf{Self-Supervised Learning}
focuses on learning visual representations by solving \emph{pretext tasks} defined on unlabeled images. Early works on self-supervised learning proposed hand-crafted pretext tasks, such as context prediction~\cite{doersch2015context}, colorization~\cite{zhang2016colorful,zhang2017split}, solving jigsaw puzzles~\cite{noroozi2016jigsaw}, predicting rotation~\cite{gidaris2018rotnet}, inpainting~\cite{pathak2016inpainting}, clustering~\cite{caron2018deepcluster}, and generative modeling~\cite{donahue2019bigbigan}.
Recent works are based on contrastive learning~\cite{gutmann2010nce,hadsell2006dimensionality}, encouraging similarity between image features under different random transformations on single input image~\cite{ye2019e2e,wu2018npid,he2019moco,misra2019pirl,chen2020simclr}.
Other approaches use contrastive losses based on context prediction~\cite{oord2018cpcv1,henaff2019cpcv2}, mutual information maximization~\cite{bachman2019amdim,hjelm2018dim,tian2019cmc}, predicting masked regions~\cite{trinh2019selfie}, and clustering~\cite{zhuang2019localagg,junnan2020pcl,caron2020swav}.

These methods lack semantic understanding as they rely on low-level visual cues (color, texture), whereas we leverage textual annotations for semantic understanding.
Unlike these methods, our approach can leverage additional metadata such as text, when scaled to internet images~\cite{sharma2018conceptual,mao2016pinterest,ordonez2011im2text}.

\noindent \textbf{Vision-and-Language Pretraining}
attempts to learn joint representations of image-text paired data that can be transferred to multimodal downstream tasks such as visual question answering~\cite{antol2015vqa,zhu2016visual7w,goyal2017making,hudson2019gqa}, visual reasoning~\cite{suhr2019corpus,zellers2019recognition}, referring expressions~\cite{kazemzadeh2014referitgame}, and language-based image retrieval~\cite{young2014image}.
Inspired by the success of BERT~\cite{devlin2019bert} in NLP,
several recent methods use Transformers~\cite{vaswani2017attention} to learn transferable joint representations of images and text~\cite{tan2019lxmert,lu2019vilbert,li2019visualbert,su2019vl,li2020unicoder,chen2019uniter,zhou2020vlp,li2020oscar}.

These methods employ complex pretraining pipelines: they typically
\textbf{(1)} start from an ImageNet-pretrained CNN;
\textbf{(2)} extract region features using an object detector fine-tuned on Visual Genome~\cite{krishna2017visual}, following \cite{anderson2018bottom};
\textbf{(3)} optionally start from a pretrained language model, such as BERT~\cite{devlin2019bert};
\textbf{(4)} combine the models from (2) and (3), and train a multimodal transformer on Conceptual Captions~\cite{sharma2018conceptual};
\textbf{(5)} fine-tune the model from (4) on the downstream task.
In this pipeline, all vision-and-language tasks are downstream from the initial visual representations learned on ImageNet.
In contrast, we pretrain via image captioning, and put vision tasks downstream from vision-and-language pretraining.

\noindent \textbf{Concurrent Work:}
Our work is closest to \emph{Sariyildiz et al.}~\cite{bulent2020icmlm} on learning visual representations from captions via image conditioned masked language modeling, with one major difference -- we train our entire model from scratch, whereas they rely on pretrained BERT for textual features. Moreover, we evaluate on additional downstream tasks like object detection and instance segmentation.
Our work is also closely related to \emph{Stroud et al.}~\cite{stroud2020learning} on learning video representations using paired textual metadata, however they solely operate and evaluate their method on video tasks.

\section{Method}
\label{sec:method}

\begin{figure*}[t]
    \centering
    \includegraphics[width=0.98\textwidth]{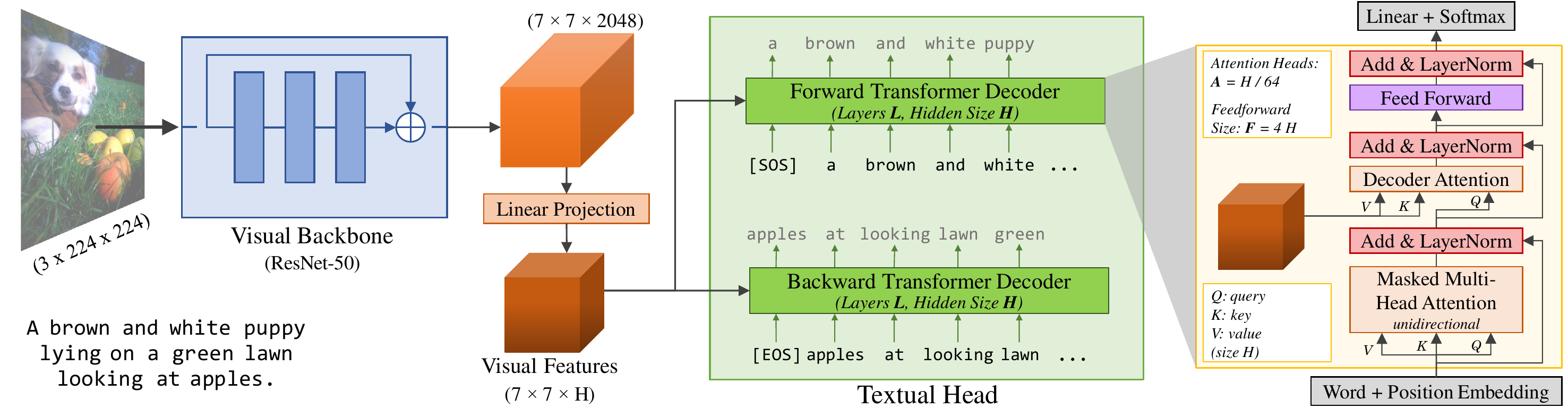}
    \caption{
        \textbf{\virtex{} pretraining setup:}
        Our model consists of a \emph{visual backbone} (ResNet-50), and a \emph{textual head} (two unidirectional Transformers).
        The visual backbone extracts image features, and textual head predicts captions via bidirectional language modeling (\emph{bicaptioning}).
        The Transformers perform masked multiheaded self-attention over caption features, and multiheaded attention over image features.
        Our model is trained end-to-end from scratch.
        After pretraining, the visual backbone is transferred to downstream visual recognition tasks.
    }
    \vspace{-10pt}
    \label{fig:system}
\end{figure*}

Given a dataset of image-caption pairs, our goal is to learn visual representations that can be transferred to downstream visual recognition tasks.
As shown in \Cref{fig:comparison},
captions carry rich semantic information about images, including the presence of objects (\emph{cat}, \emph{plate}, \emph{cake}); attributes of objects (\emph{orange and white cat}); spatial arrangement of objects (\emph{cat near a plate}); and their actions (\emph{looking at apples}).
Learned visual representations that capture such rich semantics should be useful for many downstream vision tasks.

To this end, we train \emph{image captioning} models to predict captions from images.
As shown in Figure~\ref{fig:system}, our model has two components: a \emph{visual backbone} and a \emph{textual head}.
The visual backbone extracts visual features from an input image $I$.
The textual head accepts these features and predicts a caption $C=(c_0, c_1,\ldots,c_T, c_{T+1})$ token by token,
where $c_0 = \texttt{[SOS]}$ and $c_{T+1} = \texttt{[EOS]}$ are fixed special tokens indicating the start and end of sentence.
The textual head performs bidirectional captioning (\emph{bicaptioning}):
it comprises a \emph{forward model} that predicts tokens left-to-right, and a \emph{backward model} that predicts right-to-left.
All model components are randomly initialized, and jointly trained to maximize the log-likelihood of the correct caption tokens
\vspace{-5pt}
\begin{equation}
  \begin{aligned}
    \mathcal{L}(\theta, \phi)
    = \sum_{t=1}^{T+1} \log \Big(p (c_t\mid c_{0:t-1}, I; \phi_f, \theta) \Big) \\
    + \sum_{t=0}^T \log \Big(p (c_t\mid c_{t+1:T+1}, I; \phi_b, \theta) \Big)
  \end{aligned}
\end{equation}
\vspace{-5pt}

\noindent where $\theta$, $\phi_f$, and $\phi_b$ are the parameters of the visual backbone, forward, and backward models respectively.
After training, we discard the textual head and transfer the visual backbone to downstream visual recognition tasks.

\noindent \textbf{Language Modeling:}
Our choice of pretraining task is image captioning~\cite{vinyals2015show,karpathy2015deep,donahue2015long} -- a well-studied vision-and-language task, so far kept \emph{downstream} from vision-based pretraining.
We draw inspiration from recent work in NLP using language modeling as a pretraining task to learn transferable text representations.
This involves training massive language models -- either unidirectional~\cite{peters2018deep} or bidirectional~\cite{yang2019xlnet,radford2018gpt1,radford2019gpt2,brown2020gpt3}, for predicting tokens one by one.
However, following BERT~\cite{devlin2019bert}, many large-scale models~\cite{liu2019roberta,shoeybi2019megatron} instead use \emph{masked language models} (MLMs): some tokens are randomly masked and are predicted by the model.

We performed preliminary experiments with MLMs, but like \cite{devlin2019bert,clark2020electra} we observed that MLMs converge more slowly than directional models.
We note that MLMs have poor sample efficiency, as they only predict a subset of tokens for each caption, while directional models predict all tokens.
Due to computational constraints, we focus on directional models and leave MLMs to future work.

\noindent \textbf{Visual Backbone:}
The visual backbone is a convolutional network which computes visual features of images.
It inputs raw image pixels, and outputs a spatial grid of image features.
During pretraining, these features are used to predict captions.
In downstream tasks, we either train linear models on features extracted from the visual backbone, or fine-tune the visual backbone end-to-end.

In principle we could use any convolutional network architecture for the visual backbone.
In our experiments we use a standard ResNet-50~\cite{he2016deep} as the visual backbone to facilitate comparison with our baseline methods (\Cref{sec:experiments}).
It accepts a $224\times224$ image and produces a $7\times7$ grid of $2048$-dimensional features after the final convolutional layer.
During pretraining, we apply a linear projection layer to the visual features before passing them to the textual head to facilitate decoder attention over visual features.
This projection layer is not used in downstream tasks.

\noindent \textbf{Textual Head:}
The textual head receives features from the visual backbone and predicts captions for images.
It provides a learning signal to the visual backbone during pretraining. Our overall goal is not to predict high-quality captions, but instead to learn transferable visual features.

The textual head comprises two identical language models which predict captions in forward and backward directions respectively.
Following recent advances in language modeling, we use Transformers~\cite{vaswani2017attention}, which use multiheaded self-attention both to propagate information along the sequence of caption tokens, as well as to fuse visual and textual features.
We closely follow the transformer decoder architecture from \cite{vaswani2017attention},
but use GELU~\cite{hendrycks2016gaussian} rather than ReLU, following \cite{radford2018gpt1,devlin2019bert}.
We briefly review the architecture here; refer to \cite{vaswani2017attention} for a more complete description.

During training, the forward model receives two inputs: image features from the visual backbone, and a caption describing the image.
Image features are a matrix of shape $N_I\times D_I$ giving a $D_I$-dimensional vector for each of the $N_I=7\times7$ positions in the final layer of the visual backbone.
As described earlier, the caption $C=(c_0, c_1,\ldots,c_T, c_{T+1})$ is a sequence of $T+2$ tokens, with $c_0 = \texttt{[SOS]}$ and $c_{T+1} = \texttt{[EOS]}$.
It is trained to predict $C_{1:T+1}$ token-by-token, starting with $c_0$.
The prediction $c_t$ is \emph{causal} -- it only depends on past predictions $c_{0:t-1}$ and visual features.
The backward model is similar; it operates right-to-left -- trained to predict $C_{T:0}$, given $c_{T+1}$.

First, we convert the tokens of $C$ to vectors via learned token and positional embeddings, followed by elementwise sum, layer normalization~\cite{ba2016layer} and dropout~\cite{srivastava2014dropout}.
Next, we process these vectors through a sequence of Transformer layers.
As shown in Figure~\ref{fig:system}, each layer performs masked multiheaded self-attention over token vectors, multiheaded attention between token vectors and image vectors, and applies a two-layer fully-connected network to each vector.
These three operations are each followed by dropout, wrapped in a residual connection, and followed by layer normalization.
Token vectors interact only through self-attention; the masking in this operation maintains causal structure of the final predictions.
After the last Transformer layer, we apply a linear layer to each vector to predict unnormalized log-probabilities over the token vocabulary.

The forward and backward models consist of independent Transformer layers.
However they share the same token embedding matrix (similar to \cite{peters2018deep})
which is also reused at the output layers of each model (similar to \cite{inan2016tying,press2017using}).

\noindent \textbf{Model Size:}
Several architectural hyperparameters control the size of our textual head.
We can control the \emph{width} of each Transformer layer by varying its \emph{hidden size} $H$, the number of \emph{attention heads} $A$ used in multiheaded attention, and the \emph{feedforward size} $F$ of the fully-connected network.
We follow \cite{devlin2019bert} and always set $A=H/64$ and $F=4H$; this allows us to control the width of our textual head by varying $H$.
We can also control the \emph{depth} of our textual head by varying the number of transformer layers $L$.

\noindent \textbf{Tokenization:}
We tokenize captions with SentencePiece~\cite{kudo2018sentencepiece} using the BPE algorithm~\cite{sennrich2016bpe}.
Prior to tokenization we lowercase and strip accents from captions.
We build a vocabulary of 10K tokens, including boundary (\texttt{[SOS]}, \texttt{[EOS]}) and out-of-vocab (\texttt{[UNK]}) tokens.
Following~\cite{radford2018gpt1,radford2019gpt2} we restrict subword merges between letters and punctuation to prevent redundant tokens such as \texttt{dog?} and \texttt{dog!}.
Compared to basic tokenization schemes often used for image captioning that split on whitespace~\cite{vinyals2015show,karpathy2015deep}, BPE makes fewer linguistic assumptions, exploits subword information, and results in fewer out-of-vocab tokens.

\noindent \textbf{Training Details:}
We train on the \texttt{train2017} split of the COCO Captions dataset~\cite{chen2015microsoft},
which provides $118K$ images with five captions each.
During training we apply standard data augmentation:
we randomly crop to 20-100\% of the original image size,
apply color jitter (brightness, contrast, saturation, hue), and normalize using the ImageNet mean color.
We also apply random horizontal flips, also interchanging the words `left' and `right' in the caption.

We train using SGD with momentum 0.9~\cite{polyak1964momentum,sutskever2013importance} and weight decay $10^{-4}$ wrapped in LookAhead~\cite{zhang2019lookahead} with $\alpha=0.5$ and 5 steps.
Following \cite{devlin2019bert}, we do not apply weight decay to layer normalization and bias parameters in Transformers.
We perform distributed training across 8 GPUs with batch normalization~\cite{ioffe2015batch} per GPU, following~\cite{goyal2019scaling}.
We train with a batch size of 256 images (32 per GPU) for 500K iterations ($\approx$1080 epochs). 
We use linear learning rate warmup~\cite{goyal2019scaling} for the first 10K iterations followed by cosine decay~\cite{loshchilov2016sgdr} to zero.
We found that the visual backbone required a higher LR than the textual head for faster convergence.
The visual backbone uses a max LR of $2\times10^{-1}$; the textual head uses $10^{-3}$.
We implement our models using PyTorch~\cite{pytorch} with native automatic mixed-precision~\cite{micikevicius2018mixed}.

We observe that performance on image captioning has a positive but imprecise correlation with performance on downstream visual recognition tasks (Refer Appendix \ref{subsec:best_ckpt_supp}).
We thus perform \emph{early stopping} based on the performance of our visual backbone on downstream \voc{}~\cite{everingham2009voc} linear classification (see \Cref{subsec:linear_clf}) since it is fast to evaluate and correlates well with our other downstream tasks.

\section{Experiments}
\label{sec:experiments}

In our experiments, we aim to demonstrate the effectiveness of learning visual features via natural language supervision.
As described in \Cref{sec:method}, we jointly train a \virtex{} model from scratch on the COCO Captions~\cite{chen2015microsoft} dataset.
Here, we evaluate the features learned by visual backbone on six downstream vision tasks.
We select these tasks based on two common mechanisms for transfer learning:
where the visual backbone is either used as
(a) frozen feature extractor, or
(b) weight initialization for fine-tuning.


\begin{table}[t]
    \centering \footnotesize

    \setlength{\tabcolsep}{3pt}
    \begin{tabularx}{\linewidth}{Xcccll}
    \toprule
    \textbf{Method} & \textbf{Annotations} & \textbf{Cost (hours)}$\dagger$ & \textbf{\vocclf{}} & \textbf{\inclf{}} \\
    \midrule
    \mocococo{}           & self-sup. & --                                & 63.3 & 41.1 \\
    Multi-label Clf.      & labels    & 11.1K~\cite{lin2014microsoft}     & 86.2 & 46.2 \\
    Instance Segmentation & masks     & 30.0K~\cite{lin2014microsoft}     & 82.3 & 51.0 \\
    \midrule
    \virtex{} (1 caption) & captions  & ~~~~1.3K~\cite{agrawal2019nocaps} & 84.2 & 50.2 \\
    \virtex{} (5 caption) & captions  & ~~~~6.5K~\cite{agrawal2019nocaps} & \textbf{88.7} & \textbf{53.8} \\
    \bottomrule
    \end{tabularx}
    \caption{
        \textbf{Annotation Cost Efficiency:}
        We compare downstream performance of various pretraining methods on COCO.
        \virtex{} outperforms all other methods trained on the same set of images
        with best performance vs. cost tradeoff.
        \text{\footnotesize{$\dagger$: \textit{For COCO} \texttt{train2017} \textit{split, see Appendix \ref{subsec:linearclf_appendix} for more details}.}}
    }
    \label{tab:linear_clf1}
    \vspace{-10pt}
\end{table}

\subsection{Image Classification with Linear Models}
\label{subsec:linear_clf}

Our first set of evaluations involve training linear models on frozen visual backbones --
we compare \virtex{} with various pretraining methods to test our two hypotheses:
\begin{compactenum}[\hspace{1pt}1.]
  \item Learning visual features via captions is cheaper than using other types of annotations, like labels and masks.
  \item Using semantically dense captions helps with learning effective visual features using fewer training images.
\end{compactenum}

\noindent We evaluate on two datasets:
\voc{}~\cite{everingham2009voc} and \imagenet{}~\cite{russakovsky2015imagenet}.
We choose these tasks based on their simplicity and evaluation speed.
We briefly describe the setup here.
Refer Appendix \ref{subsec:linearclf_appendix} for more details.

\noindent \textbf{\voc{}:}
We follow same protocol as SwAV~\cite{caron2020swav} (highly similar to \cite{goyal2019scaling,misra2019pirl}); we train on \vocclf{} \texttt{trainval} split (9K images, 20 classes) and report mAP on \texttt{test} split.
We train per-class SVMs on 2048-dimensional global average pooled features extracted from the last layer of the visual backbone.
For each class, we train SVMs for cost values $C \in \{0.01, 0.1, 1, 10\}$ and select best $C$ by 3-fold cross-validation. Other SVM hyperparameters are same as~\cite{goyal2019scaling}.

\noindent \textbf{\imagenet{}:}
We follow similar protocol as MoCo~\cite{he2019moco} and SwAV~\cite{caron2020swav}: we train on the ILSVRC 2012 \texttt{train} split and report top-1 accuracy on \texttt{val} split.
We train a linear classifier (fully connected layer + softmax) on 2048-dimensional global average pooled features extracted from the last layer of the visual backbone.
We train with batch size 256 distributed across 8 GPUs for 100 epochs. We use SGD with momentum $0.9$ and weight decay $0$. 
We set the initial LR to $0.3$ and decay it to zero by cosine schedule.

\begin{figure}[t]
    \centering \footnotesize
    \includegraphics[width=\linewidth]{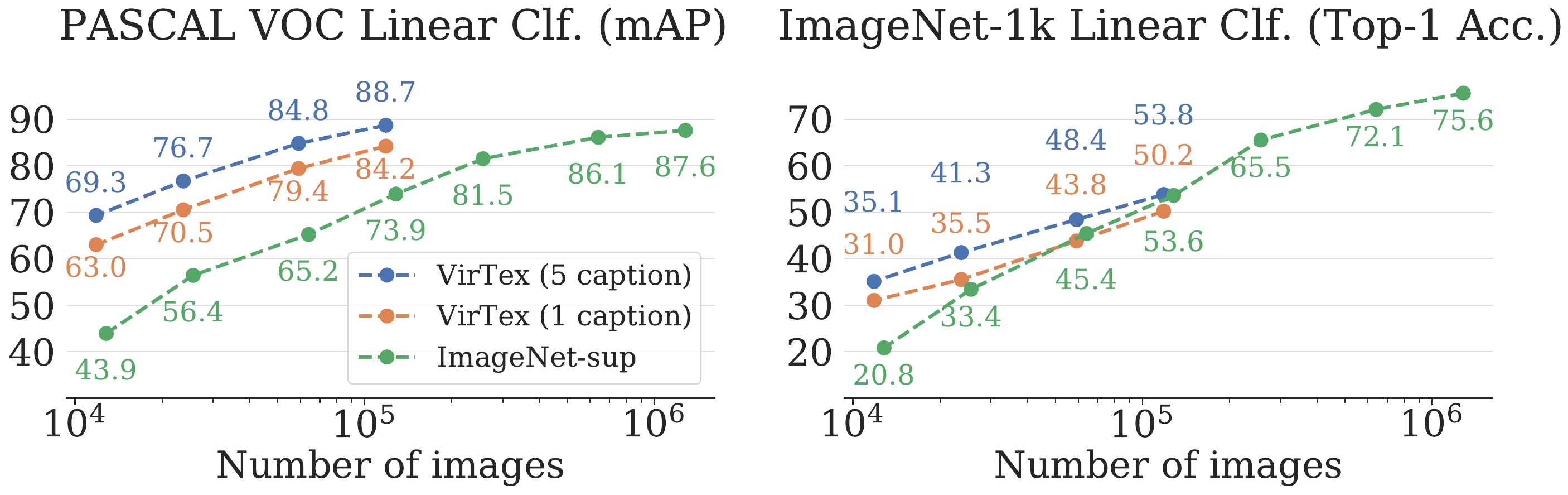}
    \caption{
        \textbf{Data Efficiency:}
        We compare \virtex{} and \insup{} models trained using varying amounts of images.
        \virtex{} closely matches or significantly outperforms \insup{} on downstream tasks despite using 10$\times$ fewer images.
        \text{\footnotesize{\textit{\inclf{}: Models using} $\le 10^5$ \emph{images are mean of 5 trials, std dev.} $\le 1.0$.}}
    }
    \label{fig:linear_clf2}
    \vspace{-10pt}
\end{figure}


\noindent \textbf{Annotation Cost Efficiency:}
We believe that using captions is appealing due to a simple and cost-efficient collection pipeline.
Here, we test our first hypothesis by comparing various pretraining methods on COCO,
each drawing supervision from different annotation types (\Cref{fig:comparison}):
\begin{compactitem}[\hspace{1pt}--]
  \item \textbf{\mocococo{} (self-supervised):} We train a MoCo-v1 model on COCO images with default hyperparameters.
  \item \textbf{Multi-label Classification (labels):} We use COCO object detection annotations (80 classes), and train a ResNet-50 backbone to predict a $K$-hot vector with values $1/K$ with a KL-divergence loss, similar to \cite{mahajan2018exploring}.
  \item \textbf{Instance Segmentation (masks):} We use a pretrained Mask R-CNN from Detectron2 model zoo~\cite{wu2019detectron2}, and extract its ResNet-50 backbone for downstream tasks. This model is trained from scratch on COCO, following \cite{he2019rethinking}.
  \item \textbf{\virtex{} (captions):} We train a \virtex{} model on COCO Captions, with ResNet-50 visual backbone and $L=1, H=2048$ textual head. Note that COCO Captions provides five captions per image, which effectively increases image-caption pairs by five-fold. Hence for a fair comparison, we also train an additional \virtex{} model using only one randomly selected caption per image.
\end{compactitem}

\noindent Results are shown in \Cref{tab:linear_clf1}.
We also compare annotation costs in terms of worker hours.
For labels and masks, we use estimates reported by COCO~\cite{lin2014microsoft}.
For captions, we estimate the cost based on \texttt{nocaps}~\cite{agrawal2019nocaps}~\footnote{We could not find estimates for COCO Captions in existing literature.}, that follows a similar data collection protocol as COCO.
We observe that \virtex{} outperforms all methods, and has the best performance vs. cost tradeoff, indicating that learning visual features using captions is more cost-efficient than labels or masks.


\noindent \textbf{Data Efficiency:}
We believe that the semantic density of captions should allow \virtex{} to learn effective visual features from fewer images than other methods.
To test our hypothesis, we compare \virtex{} and ImageNet-supervised models (\textbf{\insup{}}) trained using varying amount of images from COCO Captions and \imagenet{} respectively.

We train 4 \virtex{} models using $\{10, 20, 50, 100\}\%$ of COCO Captions (118K images) and 7 ResNet-50 models using $\{1, 2, 5, 10, 20, 50, 100\}\%$ of \imagenet{} (1.28M images).
Similar to prior experiments, we also train 4 \virtex{} models using one randomly selected caption per image.
All \virtex{} models use $L=1, H=2048$ textual heads.

We show results in \Cref{fig:linear_clf2}. On \vocclf{}, \virtex{}-$100\%$ outperforms \insup{}-$100\%$ (mAP \textbf{\color{RoyalBlue}{88.7}} vs \textbf{\color{ForestGreen}{87.6}}), despite using $10\times$ fewer images (118K vs. 1.28M).
When using similar amount of images, \virtex{} consistently outperforms \insup{} (\textbf{\color{RoyalBlue}{blue}}, \textbf{\color{RedOrange}{orange}} vs \textbf{\color{ForestGreen}{green}}), indicating superior data efficiency of \virtex{}.
We also observe that given the same number of captions for training, it is better to spread them over more images -- \virtex{}-$50\%$ (1 caption) significantly outperforms \virtex{}-$10\%$ (5 captions) (mAP \textbf{\color{RoyalBlue}{79.4}} vs \textbf{\color{RedOrange}{69.3}}).

Comparison with \insup{} on \imagenet{} classification is unfair for \virtex{}, since \insup{} models are trained for the downstream task, using the downstream dataset.
Even so, \virtex{}-$100\%$ outperforms \insupten{} (\textbf{\color{RoyalBlue}{53.8}} vs. \textbf{\color{ForestGreen}{53.6}}, 118K vs. 128K images), and consistently outperforms it when both methods use fewer than 100K images.

\begin{table}[t!]
    \centering \footnotesize
    \setlength{\tabcolsep}{2pt}
    \begin{tabularx}{\linewidth}{Xcccc}
    \toprule
    \textbf{Method} & \textbf{Pretrain Images} & \textbf{Annotations} & \textbf{\vocclf{}} & \textbf{\inclf{}} \\
    \midrule
    \mocoin{} v1~\cite{he2019moco}        & 1.28M & self-sup. & 79.4 & 60.8 \\
    PCL v1~\cite{junnan2020pcl}           & 1.28M & self-sup. & 83.1 & 61.5 \\
    SwAV (200 ep.)~\cite{caron2020swav}   & 1.28M & self-sup. & 87.9 & 72.7 \\
    \midrule
    ICMLM\textsubscript{\texttt{att-fc}}~\cite{bulent2020icmlm}~$\dagger$ & 118K & captions & 87.5 & 47.9 \\
    \virtex{}                              & 118K & captions  & 88.7 & 53.8 \\
    \bottomrule
    \end{tabularx}
    \caption{
        \textbf{Comparison with other methods:}
        We compare downstream performance of \virtex{} with recent SSL methods and concurrent work.
        \textit{\footnotesize{$\dagger$: Uses pretrained BERT-base.}}
    }
    \vspace{-10pt}
    \label{tab:linear_clf3}
\end{table}

\noindent \textbf{Comparison with other methods}:
Here, we compare \virtex{} with recent pretraining methods that have demonstrated competitive performance on downstream tasks.

\begin{compactitem}[\hspace{1pt}--]
  \item \textbf{Self-supervised pretraining:}
  We choose three recent methods based on their availability and compatibility with our evaluation setup
  -- MoCo~\cite{he2019moco}, PCL~\cite{junnan2020pcl}, and SwAV~\cite{caron2020swav}.
  We choose models trained with a similar compute budget as ours (8 GPUs, 200 ImageNet epochs).
  \item \textbf{ICMLM (Concurrent Work):}
  We adapt numbers from \emph{Sariyildiz et al.}~\cite{bulent2020icmlm}; evaluation may slightly differ.
  This model uses pretrained BERT~\cite{devlin2019bert} for textual features.
  \item \textbf{Note on vision-and-language pretraining:}
  Since we use captions, we also consider methods that learn multimodal representations for downstream vision-and-language tasks~\cite{tan2019lxmert,lu2019vilbert,li2019visualbert,su2019vl,li2020unicoder,chen2019uniter,zhou2020vlp,li2020oscar}).
  As described in \Cref{sec:related}, all these methods use an object detector trained on Visual Genome~\cite{krishna2017visual} (with ImageNet-pretrained backbone) to extract visual features, made available by \cite{anderson2018bottom}.
  These features are kept frozen, and do not learn from any textual supervision at all.
  Our comparison with ImageNet-supervised models subsumes this family of models.
\end{compactitem}

\noindent Results are shown in \Cref{tab:linear_clf3}.
\virtex{} outperforms all methods on \vocclf{}, despite being trained with much fewer images.
On \imagenet{}, comparison between self-supervised models and \virtex{} is unfair on both ends, as the former
observes downstream images during pretraining, while the latter uses annotated images.


\begin{figure}[t!]
    \centering
    \includegraphics[width=\linewidth]{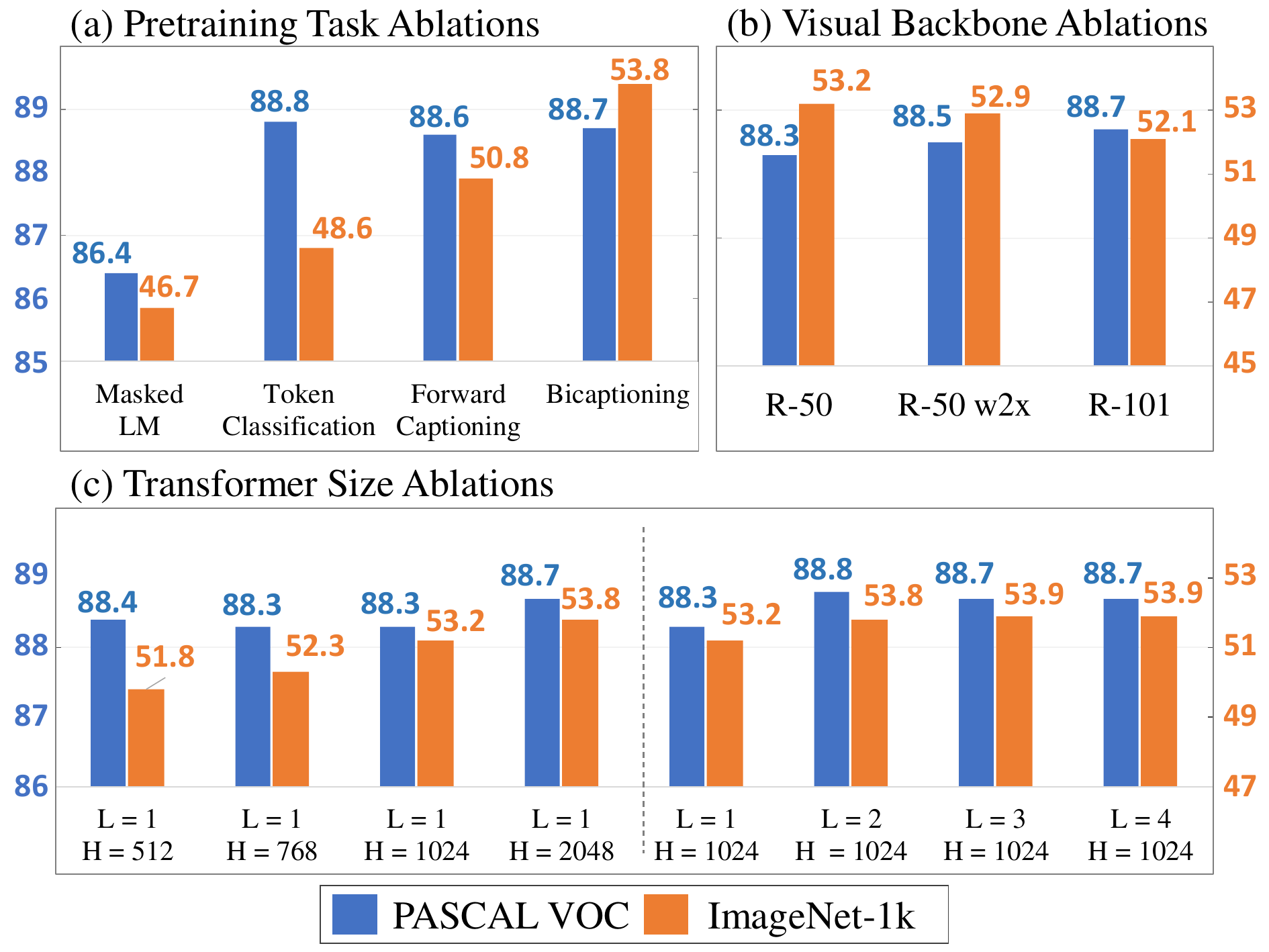}
    \caption{
        \textbf{Ablations. (a) Pretraining Tasks:} Bicaptioning improves over weaker pretraining tasks -- forward captioning, token classification and masked language modeling.
        \textbf{(b) Visual Backbone:} Bigger visual backbones improve downstream performance -- both, wider (R-50 w$2\times$) and deeper (R-101).
        \textbf{(c) Transformer Size:} Larger transformers (wider and deeper) improve downstream performance.
    }
    \vspace{-10pt}
    \label{fig:ablations}
\end{figure}

\subsection{Ablations}
\label{subsec:ablations}

The preceeding linear classification experiments demonstrate the effectiveness and data-efficiency of \virtex{}.
In this section, we conduct ablation studies to isolate the effects of our pretraining setup and modeling decisions, and uncover performance trends to seed intuition for future work.
We evaluate all ablations on \voc{} and \imagenet{} linear classification, as described in~\Cref{subsec:linear_clf}.

\begin{table*}[t]
    \newcommand{\apbbox}[1]{AP$^\text{bbox}_\text{#1}$}
    \newcommand{\apmask}[1]{AP$^\text{mask}_\text{#1}$}
    \newcolumntype{Y}{>{\raggedright\arraybackslash}X}
    \newcolumntype{Z}{>{\centering\arraybackslash}X}

    \centering
    \footnotesize
    \setlength\tabcolsep{1pt}
    \renewcommand{\arraystretch}{1.2}

    \begin{tabularx}{\textwidth}{c l  Z c YYYYYY c YYY c YYY c Y}
    \toprule
    & \multicolumn{1}{l}{\bf \multirow[b]{2}{*}{Method}}
    & \multicolumn{1}{c}{\bf \multirow[b]{2}{*}{\shortstack{Pretrain\\Images}}}
    && \multicolumn{6}{c}{\bf COCO \scriptsize{Instance Segmentation}}
    &~~~& \multicolumn{3}{c}{\bf LVIS \scriptsize{Instance Segmentation}}
    &~~~& \multicolumn{3}{c}{\bf \voc{} \scriptsize{Detection}}
    &~~~& \multicolumn{1}{c}{\bf \inatclf{}} \\
    \cmidrule{5-10} \cmidrule{12-14} \cmidrule{16-18} \cmidrule{20-20}

                &               &       &&  \apbbox{all}    & \apbbox{50}       & \apbbox{75}       &
                                            \apmask{all}    & \apmask{50}       & \apmask{75}       &&
                                            \apmask{all}    & \apmask{50}       & \apmask{75}       &&
                                            \apbbox{all}    & \apbbox{50}       & \apbbox{75}       && Top-1 \\
    \midrule
    \band
    \ttbf{1)}  & \random{}      &       &&  36.7            & 56.7              & 40.0              &
                                            33.7            & 53.8              & 35.9              &&
                                            17.4            & 27.8              & 18.4              &&
                                            33.8            & 60.2              & 33.1              && 61.4 \\

    \ttbf{2)}  & \insup{}       & 1.28M &&  41.1            & 62.0              & 44.9              &
                                            37.2            & 59.1              & 40.0              &&
                                            22.6            & 35.1              & 23.7              &&
                                            54.3            & 81.6              & 59.7              && 65.2 \\

    \ttbf{3)}  & \insupfif{}    &  640K &&  40.3\drop{0.8}  & 61.0\Drop{1.0}    & 44.0\drop{0.9}    &
                                            36.6\drop{0.6}  & 58.0\Drop{1.1}    & 39.3\drop{0.7}    &&
                                            21.2\Drop{1.4}  & 33.3\Drop{1.8}    & 22.3\Drop{1.4}    &&
                                            52.1\Drop{2.2}  & 80.4\Drop{1.2}    & 57.0\Drop{2.7}    && 63.2\Drop{2.0} \\

    \ttbf{4)}  & \insupten{}    &  128K &&  37.9\Drop{3.2}  & 58.2\Drop{3.8}    & 41.1\Drop{3.8}    &
                                            34.7\Drop{2.5}  & 55.2\Drop{3.9}    & 37.1\Drop{2.9}    &&
                                            17.5\Drop{5.1}  & 28.0\Drop{7.1}    & 18.4\Drop{5.3}    &&
                                            42.6\Drop{11.7} & 72.0\Drop{9.6}    & 43.8\Drop{15.9}   && 60.2\Drop{4.7} \\

    \midrule

    \ttbf{5)}  & \mocoin{}      & 1.28M &&  40.8\drop{0.3}  & 61.6\drop{0.4}    & 44.7\drop{0.2}    &
                                            36.9\drop{0.3}  & 58.4\drop{0.7}    & 39.7\drop{0.3}    &&
                                            22.8\rise{0.2}  & 35.4\rise{0.3}    & 24.2\Rise{0.5}    &&
                                            56.1\Rise{1.8}  & 81.5\drop{0.1}    & 62.4\Rise{0.7}    && 63.2\Drop{1.7} \\

    \ttbf{6)}  & \mocococo{}    &  118K &&  38.5\drop{0.6}  & 58.5\Drop{3.5}    & 42.0\Drop{2.9}    &
                                            35.0\Drop{2.2}  & 55.6\Drop{3.5}    & 37.5\Drop{2.5}    &&
                                            20.7\Drop{1.9}  & 32.3\Drop{2.8}    & 21.9\Drop{1.8}    &&
                                            47.6\Drop{6.7}  & 75.4\Drop{6.2}    & 51.0\Drop{8.7}    && 60.5\Drop{4.4} \\

    \midrule

    \ttbf{7)}  & \virtex{}      &  118K &&  40.9\drop{0.2}  & 61.7\drop{0.3}    & 44.8\drop{0.1}    &
                                            36.9\drop{0.3}  & 58.4\drop{0.7}    & 39.7\drop{0.3}    &&
                                            23.0\Rise{0.4}  & 35.4\Rise{0.4}    & 24.3\Rise{0.6}    &&
                                            55.3\Rise{1.0}  & 81.3\drop{0.3}    & 61.0\Rise{1.3}    && 63.4\Drop{1.4} \\
    \bottomrule
    \end{tabularx}

    \caption{
        \textbf{Fine-tuning Tasks for Transfer:}
        We compare \virtex{} with different pretraining methods across four downstream tasks.
        For each task, all methods use the same architecture. We initialize the ResNet-50 backbone weights from pretraining (except \random{}), which are then fine-tuned end-to-end.
        Performance gaps with \insup{} are shown on the side.
        On all tasks, \virtex{} significantly outperforms all methods that use similar amount of pretraining images.
        \virtex{} closely matches or exceeds ImageNet supervised and self-supervised methods, despite using $10\times$ fewer pretraining images.
    }
    \label{tab:finetuning}
    \vspace{-10pt}
\end{table*}

\noindent \textbf{Pretraining Task Ablations:}
We choose bicaptioning task as it gives a dense supervisory signal per caption.
To justify this choice, we form three pretraining tasks with \textit{sparser} supervisory signal and compare them with bicaptioning:
\begin{compactitem}[\hspace{1pt}--]
  \item \textbf{Forward Captioning:}
  We remove the backward transformer decoder and only perform left-to-right captioning.
  \item \textbf{Token Classification:}
  We replace the textual head with a linear layer and perform multi-label classification (\Cref{tab:linear_clf1}, row 2).
  We use the set of caption tokens as targets, completely ignoring the linguistic structure of captions.
  \item \textbf{Masked Language Modeling (MLM):}
  We use a single bidirectional transformer in the textual head, and perform BERT-like masked language modeling.
  We randomly mask $15\%$ of input tokens, and train the model to predict ground-truth tokens of masked positions.
\end{compactitem}
All textual heads with transformers have $L=1, H=2048$.

Results are shown in~\Cref{fig:ablations}(a).
Bicaptioning outperforms forward captioning, indicating that denser supervisory signal from bidirectional modeling is beneficial.
Bicaptioning and forward captioning both outperform token classification, demonstrating that learning to model the sequential structure of language improves visual features.

MLM performs quite worse than all three methods, possibly due to poor sample efficiency (discussed in \Cref{sec:method})
It may benefit from longer training schedules, however we leave this for future work due to computational constraints.

\noindent \textbf{Visual Backbone Ablations:}
Bigger visual backbones tend to show improvements on many vision tasks~\cite{he2016deep,he2017mask,xie2017aggregated}.
We investigate whether \virtex{} models with bigger visual backbones can improve downstream performance.
We train three \virtex{} models with $L=1, H=1024$ textual heads, and different visual backbones:
(a) ResNet-50 (default),
(b) ResNet-50 w2$\times$~\cite{zagoruyko2016wide} ($2\times$ channel width), and
(c) ResNet-101 ($2\times$ depth).
We observe that bigger visual backbones better results on \vocclf{}, however the trends are opposite on ImageNet (\Cref{fig:ablations}(b)).
We believe it to be an optimization issue. See Appendix \ref{subsec:ablations_supp} for comparison on other tasks.

\noindent \textbf{Transformer Size Ablations:}
Prior work in language modeling has shown that larger Transformers tend to learn better \emph{textual} features~\cite{radford2019gpt2,liu2019roberta,shoeybi2019megatron,brown2020gpt3}.
We investigate whether this holds for \virtex{}: do larger transformers in the textual head cause the visual backbone to learn better \emph{visual} features?
As discussed in Section~\ref{sec:method}, we may scale our textual head by increasing its \emph{width} (hidden size $H$) or its \emph{depth} (number of layers $L$).
We investigate both, training \virtex{} models with:
\begin{compactitem}[\hspace{1pt}--]
  \item Fixed $L = 1$, increasing $H\in\{512,768,1024,2048\}$.
  \item Fixed $H = 1024$, increasing $L\in\{1,2,3,4\}$.
\end{compactitem}
\noindent Results are shown in \Cref{fig:ablations}(c) -- increasing transformer size, both width and depth, generally improves downstream performance.
Performance degrades slightly with very deep transformers ($L = 4$), indicating overfitting.
We hope that massive transformers with billions of parameters will help when scaling \virtex{} to large-scale, more noisy image-text paired datasets~\cite{sharma2018conceptual,mao2016pinterest,ordonez2011im2text} that are larger than COCO Captions.


\begin{figure*}[t!]
    \newcolumntype{Y}{>{\centering\arraybackslash}X}
    \setlength{\tabcolsep}{1.8pt}
    \footnotesize
    \renewcommand{\arraystretch}{0.96}
    \begin{tabularx}{0.29\linewidth}{Xcccc}
        \toprule
        \textbf{Backbone} & \textbf{Depth} & \textbf{Width} & \textbf{CIDEr} & \textbf{SPICE} \\
        \midrule
        R-50 & 1 & 512  & 103.2 & 19.3 \\
        R-50 & 1 & 768  & 103.7 & 19.6 \\
        R-50 & 1 & 1024 & 103.5 & 19.8 \\
        R-50 & 1 & 2048 & \textbf{104.2} & \textbf{19.9} \\
        \midrule
        R-50 & 1 & 1024 & 103.5 & 19.8 \\
        R-50 & 2 & 1024 & \textbf{106.9} & \textbf{20.0} \\
        R-50 & 3 & 1024 & 104.3 & 19.5 \\
        R-50 & 4 & 1024 & 103.8 & 19.2 \\
        \midrule
        R-50 w2$\times$ & 1 & 1024 & 102.7 & 19.6 \\ 
        R-101 & 1 & 1024 & \textbf{106.6} & \textbf{20.1} \\
        \bottomrule
    \end{tabularx}
    \hfill
    \small
    \begin{tabularx}{0.7\linewidth}{YcYcYcY}
        \multicolumn{7}{c}{\virtex{} predicted captions (R-50, $L=1, H=512$), forward transformer decoder} \\
        \includegraphics[width=0.162\textwidth]{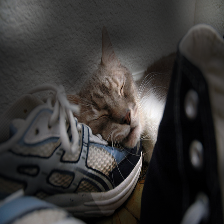} &~~&
        \includegraphics[width=0.162\textwidth]{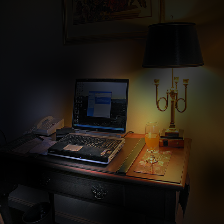} &~~&
        \includegraphics[width=0.162\textwidth]{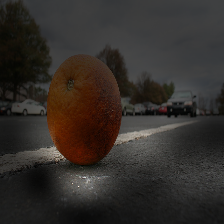} &~~&
        \includegraphics[width=0.162\textwidth]{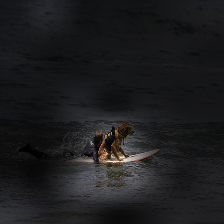} \\
        \texttt{a cat laying on a pair of blue \attention{shoes}} &&
        \texttt{a laptop computer sitting on top of a \attention{desk}} &&
        \texttt{an orange is sitting on the side of a \attention{road}} &&
        \texttt{a dog \attention{riding} on a surfboard in the ocean} \\
    \end{tabularx}
    \vspace{-5pt}
    \caption{
        \textbf{Image Captioning:}
        We report image captioning performance (\textbf{CIDEr}~\cite{vedantam2015cider} and \textbf{SPICE}~\cite{anderson2016spice})
        of \virtex{} models on COCO \texttt{val2017} split -- all variants show modest performance.
        We also show some predicted captions on the right.
        For the \attention{highlighted word}, we visualize decoder attention weights from the textual head on the input image.
        Our model focuses on relevant image regions to predict objects (\attention{shoes}, \attention{desk}),
        background (\attention{road}) as well as actions (\attention{riding}).
    }
    \label{fig:captioning}
    \vspace{-10pt}
\end{figure*}

\subsection{Fine-tuning Tasks for Transfer}
\label{subsec:finetune}

So far we have evaluated \virtex{} using features extracted from \emph{frozen} visual backbones.
Another common mechanisms for transfer learning is \emph{fine-tuning}, where the entire visual backbone is updated for the downstream task.

We evaluate features learned using \virtex{} on four downstream tasks with fine-tuning: (a) Instance Segmentation on COCO~\cite{lin2014microsoft}; (b) Instance Segmentation on LVIS~\cite{gupta2019lvis}; and (c) Object Detection on \voc{}~\cite{everingham2009voc}; (d) Fine-grained Classification on \inat{}~\cite{van2018inaturalist}.
In all these experiments, we use the \virtex{} model with ResNet-50 visual backbone and a textual head with $L=1, H=2048$.

\noindent \textbf{Baselines:}
Our main baselines are ImageNet-supervised (\insup{}) and MoCo.
We consider three variants of \insup{} pretrained with $\{10,50,100\}\%$ of ImageNet images (\Cref{fig:linear_clf2}).
Similarly for MoCo, we consider both \mocoin{} (\Cref{tab:linear_clf3}) and \mocococo{} (\Cref{tab:linear_clf1}).
We also include \random{} baseline, trained from scratch on downstream task.

We follow the same evaluation protocol as MoCo~\cite{he2019moco} for all four tasks.
We use Detectron2~\cite{wu2019detectron2} for tasks (a,b,c). Our \insup{}-$100\%$ results are slightly better than those reported in~\cite{he2019moco} -- we use pretrained ResNet-50 model from \texttt{torchvision}, whereas they used the MSRA ResNet-50 model from Detectron~\cite{girshick2018detectron}.
We briefly describe implementation details that differ from default Detectron2 settings.
Refer Appendix \ref{subsec:finetune_supp} for full details.

\noindent \textbf{COCO Instance Segmentation:}
We train Mask R-CNN~\cite{he2017mask} models with ResNet-50-FPN backbones~\cite{lin2017feature}.
We initialize backbone with pretrained weights, train on \texttt{train2017} split, and evaluate on \texttt{val2017} split.
We fine-tune all layers end-to-end with BN layers synchronized across GPUs~\cite{peng2018megdet} (\emph{SyncBN}).
We also use SyncBN in FPN layers.
We train with batch size 16 distributed across 8 GPUs, following $2\times$ schedule (180K iterations with initial LR 0.02, multiplied by 0.1 at iterations 120K and 160K).

\noindent \textbf{LVIS Instance Segmentation:}
The LVIS dataset provides instance segmentation labels for a long tail of 1203 entry-level object categories, and stresses the ability to recognize many object types from few training samples.
We train Mask R-CNN models with ResNet-50-FPN backbones on \texttt{train\_v1.0} and evaluate on \texttt{val\_v1.0} split.
Following MoCo settings, we keep BN parameters frozen for all \insup{} baselines.
We train with $2\times$ schedule as COCO, use class resampling and test-time hyperparameters (0.0 score threshold and 300 detections per image) same as \cite{gupta2019lvis}.

\noindent \textbf{\voc{} Detection:}
We train Faster R-CNN~\cite{ren2015faster} models with ResNet-50-C4 backbones on \texttt{trainval07+12} split, and evaluate on \texttt{test2007} split. Like COCO, we fine-tune all models with SyncBN. We train for 24K iterations, including linear LR warmup for first 100 iterations. We set the maximum LR as 0.02, that is divided by 10 at iterations 18K and 22K. 
We distribute training across 8 GPUs, with batch size 2 per GPU.
We use gradient checkpointing~\cite{martens2012training,chen2016training} to reduce the heavy memory footprint of these models and train them with desired batch size on our 12 GB GPUs.

\noindent \textbf{\inat{} Fine-grained Classification:}
The \inat{} dataset provides labeled images for 8142 fine-grained categories, with a long-tailed distribution.
We fine-tune the pretrained ResNet-50 with a linear layer end-to-end.
We train on \texttt{train2018} split and evaluate on \texttt{val2018} split, following training setup from \texttt{torchvision} --
we train for 100 epochs using SGD with momentum 0.9 and weight decay $10^{-4}$, and batch size 256 distributed across 8 GPUs. Fine-tuning uses LR 0.025 (and \random{} uses 0.1), which is multiplied by 0.1 at epochs 70 and 90.

\noindent \textbf{Results:} We show results in \Cref{tab:finetuning}.
\virtex{} matches or exceeds ImageNet-supervised pretraining and \mocoin{} on all tasks (row \ttbf{2,5} vs. \ttbf{7}) despite using 10$\times$ fewer pretraining images.
Moreover, \virtex{} significantly outperforms methods that use similar, or more pretraining images (row \ttbf{3,4,6} vs. \ttbf{7}), indicating its superior data-efficiency.
Among all tasks, \virtex{} shows significant improvements on LVIS, that shows the effectiveness of natural language annotations in capturing the long tail of visual concepts in the real world.


\subsection{Image Captioning}
\label{subsec:captioning}

Our goal is to learn transferable visual features via textual supervision.
To do so, we use image captioning as a pretraining task.
Although our goal is not to advance the state-of-the-art in image captioning,
in Figure~\ref{fig:captioning} we show quantitative and qualitative results of \virtex{} models trained from scratch on COCO.
All models show modest performance, far from current state-of-the-art methods, that commonly involve some pretraining.
However, captioning metrics are known to correlate weakly with human judgement -- we surpass human performance on COCO.

We show some predicted captions by \virtex{} (R-50, $L = 1, H = 512$) model.
We apply \emph{beam search} on the forward transformer decoder (5 beams) to decode most likely captions.
The \emph{decoder attention module} in this transformer attends over a $7 \times 7$ grid of image features through $A = 8$ heads at each time-step for predicting a token.
We average these $7 \times 7$ attention weights over all the heads, and overlay them on $224 \times 224$ input image (via bicubic upsampling). 

In Figure~\ref{fig:captioning}, we show visualizations for some tokens. We observe that our model attends to relevant image regions
for making predictions, indicating that \virtex{} learns meaningful visual features with good semantic understanding.

\section{Conclusion}
\label{sec:conclusion}

We have shown that learning visual representations using textual annotations can be competitive to methods based on supervised classification and self-supervised learning on ImageNet.
We solely focus on downstream vision tasks -- future works can explore other tasks that transfer both the visual backbone and the textual head.
Finally, the usage of captions opens a clear pathway to scaling our approach to web-scale image-text pairs, that are orders of magnitude larger, albeit more noisy than COCO Captions.

\clearpage
\section*{Acknowledgments}

We thank Harsh Agrawal, Mohamed El Banani, Richard Higgins, Nilesh Kulkarni and Chris Rockwell for helpful discussions and feedback on the paper.
We thank Ishan Misra for discussions regarding PIRL/SwAV evaluation protocol;
Saining Xie for discussions about replicating iNaturalist evaluation as MoCo;
Ross Girshick and Yuxin Wu for help with Detectron2 model zoo;
Georgia Gkioxari for suggesting the Instance Segmentation pretraining task ablation;
and Stefan Lee for suggestions on figure aesthetics.
We thank Jia Deng for access to extra GPUs during project development;
and UMich ARC-TS team for support with GPU cluster management.
Finally, we thank all the Starbucks outlets in Ann Arbor for many hours of free WiFi.
This work was partially supported by the Toyota Research Institute (TRI).
However, note that this article solely reflects the opinions and conclusions of its authors and not TRI or any other Toyota entity.

{\small
\bibliographystyle{ieeetr_fullname}
\bibliography{references.bib}
}

\clearpage

\begin{appendices}
   \section{Additional Experiments}
\label{sec:experiments_supp}

In this section, we describe additional implementation details about our experiments in \Cref{sec:experiments}.
Our evaluation protocol is consistent with prior works on pretraining visual representations -- we report differences where applicable.


\subsection{Image Classification with Linear Models}
\label{subsec:linearclf_appendix}

\noindent \textbf{\voc{}:}
We use standard data augmentation on images from both \texttt{trainval} and \texttt{test} split --
we resize the shorter edge to 256 pixels, and take a $224 \times 224$ center crop.
We normalize images by ImageNet color (RGB mean = $[0.485, 0.456, 0.406]$, std = $[0.229, 0.224, 0.225]$).

Prior works~\cite{goyal2019scaling,misra2019pirl,caron2020swav} train per-class SVMs for $C \in [2^{-19}, 2^{-4}] \cup [10^{-7}, 10^{-2}]$ (26 values), and choose best SVM based on 3-fold cross-validation.
In our initial evaluations, we observed that the best performing SVMs are typically trained with cost values $C \in \{0.01, 0.1, 1.0, 10.0 \}$.
Based on this observation, we only use these values for faster evaluation.
For training SVMs, we use scikit-learn~\cite{scikit-learn} with LIBLINEAR~\cite{fan2008liblinear} backend, default parameters are:
\textcolor{RoyalBlue}{\texttt{LinearSVC(penalty=`l2', dual=True,\\
max\_iter=2000, tol=1e-4, class\_weight=\{1: 2, -1: 1\},
loss=`squared\_hinge')}}.

\noindent \textbf{\imagenet{}:}
For data augmentation during training, we randomly crop 20--100$\%$ of the original image size, with a random aspect ratio in $(4/3, 3/4)$, resize to $224\times224$, apply random flip, and normalization by ImageNet color.
During evaluation, we resize the shorter edge to 256 pixels and take a $224\times224$ center crop.
We initialize the weights of the linear layer as $N(0.0, 0.01)$, and bias values as 0.

Note that we perform a small LR sweep separately for our \virtex{} model (ResNet-50 and $L=1, H=2048$), and ImageNet-supervised models. For \Cref{fig:linear_clf2}, best LR values for \virtex{} models is 0.3 (as mentioned in \Cref{subsec:linear_clf}, and ImageNet-supervised models is 0.1.

\noindent \textbf{Annotation Cost Efficiency:}
Here, we provide details on our cost estimates for different methods in \Cref{tab:linear_clf1}.
For labels and masks, we use estimates reported by COCO~\cite{lin2014microsoft}, and for captions we use estimates reported by \texttt{nocaps}~\cite{agrawal2019nocaps}, collected in a similar fashion as COCO.
\begin{compactitem}[\hspace{1pt}--]
    \item \textbf{Labels:} We consider total time of \emph{Category Labeling} and \emph{Instance Spotting} steps in \cite{lin2014microsoft} ($\sim$30K hours).
    This estimate corresponds to 328K images -- we scale it for COCO Captions \texttt{train2017} split (118K images).
    \item \textbf{Masks:}
    As reported in \cite{lin2014microsoft}, it takes 22 worker hours for collecting 1000 instance segmentation masks.
    We use this estimate to compute time for $\sim$860K masks in COCO \texttt{train2017} split.
    The collection of masks is dependent on \emph{Category Labeling} and \emph{Instance Spotting}, we add the time for collecting labels in our total estimate.
    \item \textbf{Captions:}
    We use the median time per caption (39.2 seconds) as reported in \cite{agrawal2019nocaps} ($\sim$151K captions) to estimate the cost of collecting (118K $\times 5$) captions in COCO.
\end{compactitem}

\noindent \textbf{Data Efficiency:}
We train our ImageNet-supervised models on randomly sampled subsets of ImageNet ($1\%$, $2\%$ $5\%$, $10\%$, $20\%$, $50\%$).
We sample training examples such that the class distribution remains close to $100\%$ ImageNet.
For \virtex{} models, we randomly sample $10\%$, $20\%$, $50\%$, and $100\%$ of COCO Captions~\cite{chen2015microsoft} --
we do not use any class labels to enforce uniform class distribution. Note that \emph{this may put ImageNet-supervised models at an advantage}.

We train our ImageNet-supervised models by following the \emph{exact} setup used to train the publicly available ResNet-50 model in \texttt{torchvision}.
We use SGD with momentum 0.9 and weight decay $10^{-4}$.
We use a batch size of 256, and perform distributed training across 8 GPUs (batch size 32 per GPU).
We train for 90 epochs, with an initial learning rate 0.1, that is divided by 10 at epochs 30 and 60.
We keep the number of training epochs fixed for models trained on smaller subsets of ImageNet (else they tend to overfit).
For \virtex{} models, we scale training iterations according to the size of the sampled training set.

\noindent \textbf{Comparison: ImageNet vs. Cropped COCO.}
Note that the ImageNet images mostly contain a single object (commonly called \emph{iconic} images).
On the other hand, COCO dataset contains $\sim$2.9 object classes and $\sim$5.7 instances per image.
It may seem that \virtex{} requires fewer images than ImageNet-supervised models as they contain multiple objects per image.
Here, we make an additional comparison to control the varying image statistics between datasets.

Specifically, we crop objects from COCO images and create a dataset of 860K \emph{iconic} images.
We randomly expand bounding boxes on all edges by 0--30 pixels before cropping, to mimic ImageNet-like images.
We train a ResNet-50 with same hyperparameters as ImageNet-supervised models, described above.
It achieves \textbf{79.1} \vocclf{} mAP (vs. \textbf{88.7} \virtex{}).
This shows that the data-efficiency of \virtex{} does not \emph{entirely} stem from using scene images with multiple objects.

\begin{figure}[t]
    \centering \footnotesize
    \includegraphics[width=\linewidth]{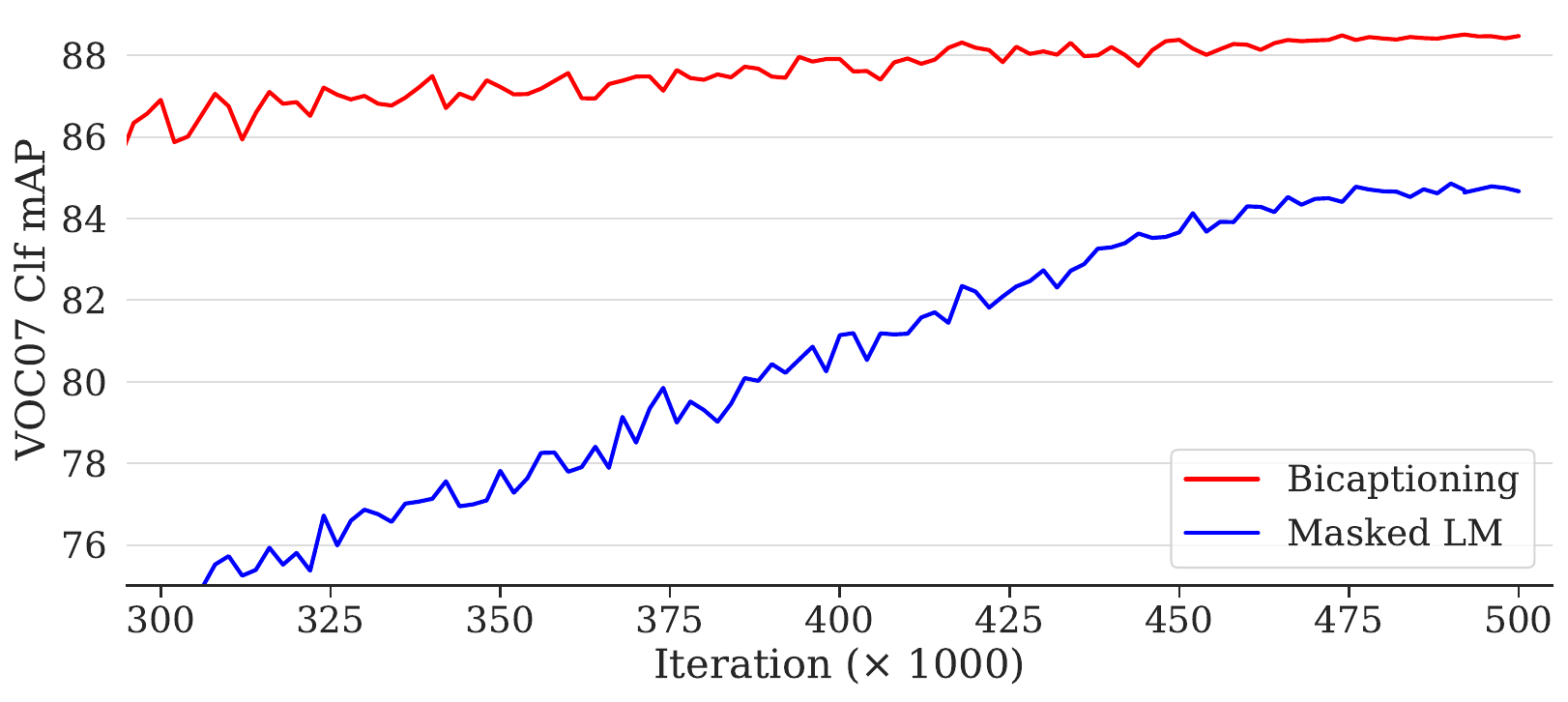}
    \caption{
        \textbf{Bicaptioning vs. Masked Language Modeling:}
        We compare \vocclf{} mAP of Bicaptioning and Masked LM pretraining tasks.
        We observe that Masked LM converges slower than Bicaptioning, indicating poor sample efficiency.
    }
    \label{fig:bicap_mlm}
\end{figure}


\subsection{Ablations}
\label{subsec:ablations_supp}

\begin{table}[t]
    \newcommand{\apbbox}[1]{AP$^\text{bbox}_\text{#1}$}
    \newcommand{\apmask}[1]{AP$^\text{mask}_\text{#1}$}
    \newcolumntype{Y}{>{\raggedright\arraybackslash}X}
    \centering \footnotesize \setlength\tabcolsep{1pt}

    \begin{tabularx}{\linewidth}{l c YY c YYY}
    \toprule
    \multicolumn{1}{l}{\bf \multirow[b]{2}{*}{Backbone}}
    &~~~& \textbf{\vocclf{}} & \textbf{\inclf{}}
    &~~~& \multicolumn{3}{c}{\textbf{\voc{} \scriptsize{Detection}}} \\
    \cmidrule{6-8}
    && mAP & Top-1 && \apbbox{all}    & \apbbox{50}       & \apbbox{75} \\
    \midrule

    \band
    ResNet-50            && 88.3 & 53.2 &&  55.2 & 81.2 & 60.8 \\
    ResNet-50 w2$\times$ && 88.5\rise{0.2} & 52.9\drop{0.3} && 56.6\Rise{1.4} & 82.0\Rise{0.8} & 62.8\Rise{2.0} \\
    ResNet-101           && 88.7\Rise{0.4} & 52.0\Drop{1.2} && 57.9\Rise{2.7} & 82.0\Rise{0.8} & 63.6\Rise{2.8} \\

    \bottomrule
    \end{tabularx}
    \caption{
        \textbf{Additional Evaluations for Backbone Ablations.}
        We compare \virtex{} models ($L=1, H=1024$) with different visual backbones.
        We observe that larger backbones generally improve downstream performance.
    }
    \label{tab:backbone_ablations_supp}
    \vspace{-10pt}
\end{table}

\noindent \textbf{Bicaptioning vs. Masked Language Modeling.}
In our pretraining task ablations (\Cref{subsec:ablations}), we observed that Masked Language Modeling performs quite worse than all other pretraining tasks on downstream linear classification performance.
This issue arises from the poor sample efficiency of Masked LM, discussed in \Cref{sec:method}.

For more evidence, we inspect \vocclf{} mAP of Masked LM, validated periodically during training.
In \Cref{fig:bicap_mlm}, we compare this with \vocclf{} mAP of Bicaptioning. Both models use $L = 1, H = 2048$ textual heads.
We find that Masked LM indeed converges slower than bicaptioning, as it receives weaker supervision per training caption -- only corresponding to masked tokens.
We believe that a longer training schedule may lead to MLM outperforming bicaptioning, based on its success in language pretraining~\cite{devlin2019bert}.

\noindent \textbf{Additional Evaluation: Backbone Ablations.}
In our backbone ablations (\Cref{fig:ablations}), we observed that larger visual backbones improve \vocclf{} classification performance.
However, the performance trend for \imagenet{} linear classification is opposite.
We think this is an optimization issue -- the hyperparameters chosen for ResNet-50 may not be optimal for other backbones.
To verify our claims, we evaluate these models on \voc{} object detection.

In \Cref{tab:backbone_ablations_supp}, we observe that the performance trends of \voc{} object detection match with \vocclf{} classification.
Hence, we conclude that using larger visual backbones can improve downstream performance.


\subsection{Fine-tuning Tasks for Transfer}
\label{subsec:finetune_supp}

We described the main details for downstream fine-tuning tasks in \Cref{subsec:finetune}.
We provide config files in Detectron2~\cite{wu2019detectron2} format to exactly replicate our downstream fine-tuning setup for COCO (\Cref{tab:coco_config}), \voc{} (\Cref{tab:voc_config}), LVIS (\Cref{tab:lvis_config}).
We apply modified hyperparameters on top of base config files available at:\\
\texttt{\hyperlink{https://github.com/facebookresearch/detectron2/blob/b267c6f314f4fa86eed6818ca7611f79d40bc8e8/configs}{github.com/facebookresearch/detectron2 @ b267c6}}

\noindent \textbf{\inat{} Fine-grained Classification:}
We use data augmentation and weight initialization same as \imagenet{} linear classification (\Cref{subsec:linearclf_appendix}).
Despite a long-tailed distribution like LVIS, we do not perform class balanced resampling, following the evaluation setup of MoCo~\cite{he2019moco}.

\begin{table}
\begin{lstlisting}[language=yaml,basicstyle=\linespread{0.975}\footnotesize\ttfamily]
_BASE_: "Base-RCNN-FPN.yaml"
INPUT:
  FORMAT: "RGB"
DATASETS:
  TRAIN: ("coco_2017_train",)
  TEST: ("coco_2017_val",)
MODEL:
  WEIGHTS: "Loaded externally"
  MASK_ON: True
  PIXEL_MEAN: [123.675, 116.280, 103.530]
  PIXEL_STD: [58.395, 57.120, 57.375]
  BACKBONE:
    FREEZE_AT: 0
  RESNETS:
    DEPTH: 50
    NORM: "SyncBN"
    STRIDE_IN_1X1: False
  FPN:
    NORM: "SyncBN"
SOLVER:
  IMS_PER_BATCH: 16
  BASE_LR: 0.02
  STEPS: (120000, 160000)
  MAX_ITER: 180000
TEST:
  PRECISE_BN:
    ENABLED: True
\end{lstlisting}
    \vspace{-10pt}
    \caption{\textbf{COCO Instance Segmentation:} Detectron2 config parameters that differ from base config file.}
    \label{tab:coco_config}
    \vspace{-10pt}
\end{table}

\begin{table}
\begin{lstlisting}[language=yaml,basicstyle=\linespread{0.975}\footnotesize\ttfamily]
_BASE_: "Base-RCNN-C4.yaml"
INPUT:
  FORMAT: "RGB"
  MIN_SIZE_TRAIN: (480, 512, 544, 576, 608, 640,
                   672, 704, 736, 768, 800)
DATASETS:
  TRAIN:("voc_2007_trainval","voc_2012_trainval")
  TEST: ("voc_2007_test",)
MODEL:
  MASK_ON: False
  WEIGHTS: "Loaded externally"
  PIXEL_MEAN: [123.675, 116.280, 103.530]
  PIXEL_STD: [58.395, 57.120, 57.375]
  BACKBONE:
    FREEZE_AT: 0
  RESNETS:
    DEPTH: 50
    NORM: "SyncBN"
    STRIDE_IN_1X1: False
  FPN:
    NORM: "SyncBN"
  ROI_HEADS:
    NUM_CLASSES: 20
SOLVER:
  IMS_PER_BATCH: 16
  BASE_LR: 0.02
  STEPS: (18000, 22000)
  MAX_ITER: 24000
  WARMUP_ITERS: 100
TEST:
  PRECISE_BN:
    ENABLED: True
\end{lstlisting}
    \vspace{-10pt}
    \caption{
        \textbf{\voc{} Object Detection:} Detectron2 config parameters that differ from base config file.
    }
    \label{tab:voc_config}
    \vspace{-10pt}
\end{table}

\begin{table}
\begin{lstlisting}[language=yaml,basicstyle=\linespread{0.975}\footnotesize\ttfamily]
_BASE_: "Base-RCNN-FPN.yaml"
INPUT:
  FORMAT: "RGB"
DATASETS:
  TRAIN: ("lvis_v1_train",)
  TEST: ("lvis_v1_val",)
DATALOADER:
  SAMPLER_TRAIN: "RepeatFactorTrainingSampler"
  REPEAT_THRESHOLD: 0.001
MODEL:
  WEIGHTS: "Loaded externally"
  MASK_ON: True
  PIXEL_MEAN: [123.675, 116.280, 103.530]
  PIXEL_STD: [58.395, 57.120, 57.375]
  BACKBONE:
    FREEZE_AT: 0
  RESNETS:
    DEPTH: 50
    NORM: "SyncBN"  # For IN-sup: "FrozenBN"
    STRIDE_IN_1X1: False
  FPN:
    NORM: "SyncBN"  # For IN-sup: ""
  ROI_HEADS:
    NUM_CLASSES: 1203
    SCORE_THRESH_TEST: 0.0001
SOLVER:
  IMS_PER_BATCH: 16
  BASE_LR: 0.02
  STEPS: (120000, 160000)
  MAX_ITER: 180000
TEST:
  DETECTIONS_PER_IMAGE: 300
  PRECISE_BN:
    ENABLED: True
\end{lstlisting}
    \vspace{-10pt}
    \caption{
        \textbf{LVIS Instance Segmentation:} Detectron2 config parameters that differ from base config file.
    }
    \label{tab:lvis_config}
\end{table}

\noindent \textbf{LVIS v0.5 Instance Segmentation:}
In \Cref{subsec:finetune}, we evaluated \virtex{} and baseline methods on LVIS Instance Segmentation task using LVIS \texttt{v1.0 train} and \texttt{val} splits.
One of our baselines, MoCo, conducted this evaluation using \texttt{LVIS v0.5} splits.
For completeness, we report additional results on LVIS \texttt{v0.5} split.
The main changes in config (\Cref{tab:lvis_config}) following original LVIS \texttt{v0.5} baselines are: \texttt{NUM\_CLASSES:} \textcolor{RoyalBlue}{\texttt{1230}} and \texttt{SCORE\_THRESHOLD\_TEST: \textcolor{RoyalBlue}{0.0}}

Results are shown in \Cref{tab:lvis5_supp}. We observe the \virtex{} significantly outperforms all baseline methods on LVIS \texttt{v0.5} split, similar to evaluation on LVIS \texttt{v1.0} split.

\begin{table}[t]
    \newcommand{\apbbox}[1]{AP$^\text{bbox}_\text{#1}$}
    \newcommand{\apmask}[1]{AP$^\text{mask}_\text{#1}$}
    \newcolumntype{Y}{>{\raggedright\arraybackslash}X}
    \newcolumntype{Z}{>{\centering\arraybackslash}X}

    \centering
    \footnotesize
    \setlength\tabcolsep{1pt}
    \renewcommand{\arraystretch}{1.1}

    \begin{tabularx}{\linewidth}{c l Z c YYY}
    \toprule
    & \multicolumn{1}{l}{\bf \multirow[b]{2}{*}{Method}}
    & \multicolumn{1}{c}{\bf \multirow[b]{2}{*}{\shortstack{Pretrain\\Images}}}
    &~~~& \multicolumn{3}{c}{\bf LVIS v0.5 \scriptsize{Instance Segmentation}} \\
    \cmidrule{5-7}

    & & & &  \apbbox{all}    & \apbbox{50}       & \apbbox{75} \\
    \midrule
    \band
    \ttbf{1)}  & \random{}      &       &&  22.5 & 34.8 & 23.8 \\
    \ttbf{2)}  & \insup{}       & 1.28M &&  24.5 & 38.0 & 26.1 \\
    \ttbf{3)}  & \insupfif{}    &  640K &&  23.7\drop{0.8} & 36.7\Drop{1.3} & 25.1\Drop{1.0} \\
    \ttbf{4)}  & \insupten{}    &  128K &&  20.5\Drop{4.0} & 32.8\Drop{6.2} & 21.7\Drop{5.2} \\
    \midrule
    \ttbf{5)}  & \mocoin{}      & 1.28M &&  24.1\drop{0.4} & 37.4\drop{0.6} & 25.5\drop{0.6} \\
    \ttbf{6)}  & \mocococo{}    &  118K &&  23.1\Drop{1.4} & 35.3\Drop{2.7} & 24.9\Drop{1.2} \\
    \midrule
    \ttbf{7)}  & \virtex{}      &  118K &&  25.4\Rise{0.9} & 39.0\Rise{1.0} & 26.9\Rise{0.8} \\
    \bottomrule
    \end{tabularx}

    \caption{
        \textbf{Downstream Evaluation: LVIS v0.5 Instance Segmentation.}
        We compare \virtex{} with different pretraining methods for LVIS \texttt{v0.5} Instance Segmentation. All methods use Mask R-CNN with ResNet-50-FPN backbone.
        Performance gaps with \insup{} are shown on the side.
        The trends are similar to LVIS \texttt{v1.0}~\Cref{tab:finetuning} -- \virtex{} significantly outperforms all baseline methods.
    }
    \label{tab:lvis5_supp}
\end{table}


\subsection{Selecting Best Checkpoint by \vocclf{} mAP}
\label{subsec:best_ckpt_supp}

As described in \Cref{sec:method}, we observed that image captioning performance has an imprecise correlation with performance on downstream vision tasks.
Hence, we select our best checkpoint based on \vocclf{} classification mAP.

\begin{figure}[t]
    \centering \footnotesize
    \includegraphics[width=\linewidth]{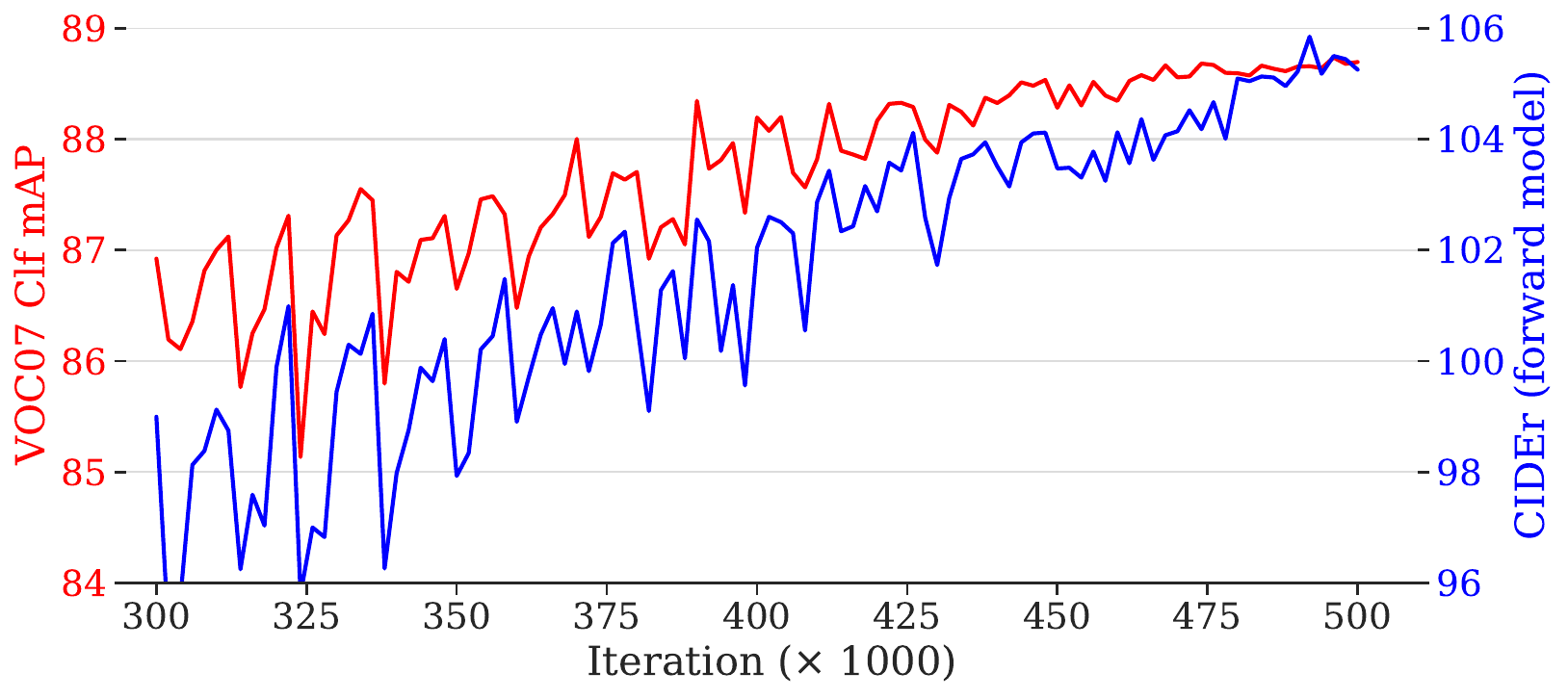}
    \caption{
        \textbf{Validation metrics: \vocclf{} mAP and CIDEr.}
        We compare \vocclf{} mAP and CIDEr score of \virtex{} (ResNet-50, $L = 1, H = 2048$) model.
        We observe that captioning performance has a positive, yet imprecise correlation with downstream performance on vision tasks.
    }
    \label{fig:voc07_cider}
\end{figure}

In \Cref{fig:voc07_cider}, we compare validation metrics of our best \virtex{} model (ResNet-50, $L = 1, H = 2048$). We observe the trends of \vocclf{} mAP and CIDEr~\cite{vedantam2015cider} score of the forward transformer decoder.
We observe that an improvement in captioning performance indicates an improvement in downstream performance.
However these are not strongly correlated --
the best performing checkpoints according to these metrics occur at different iterations: 496K according to \vocclf{} mAP (\textbf{\textcolor{Red}{88.7}}), and 492K according to CIDEr (\textbf{\textcolor{RoyalBlue}{105.8}}).
Hence, we select the best checkpoint based on \voc{} linear classification performance.
We use this task as a representative downstream vision task for evaluation due to its speed and simplicity.

\section{Decoder Attention Visualizations for Caption Predictions}
\label{sec:captioning_supp}

\begin{figure*}
  \newcolumntype{Y}{>{\centering\arraybackslash}X}
  \centering\ttfamily\footnotesize
  \setlength\tabcolsep{1pt}
  \renewcommand{\arraystretch}{1.2}
  \begin{tabularx}{\textwidth}{YYYYYYYYY}
    \includegraphics[width=0.105\textwidth]{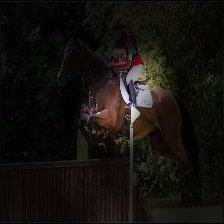} &
    \includegraphics[width=0.105\textwidth]{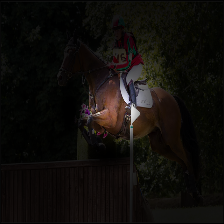} &
    \includegraphics[width=0.105\textwidth]{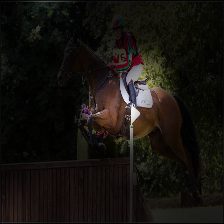} &
    \includegraphics[width=0.105\textwidth]{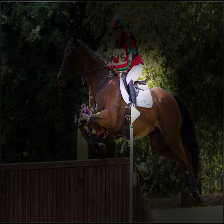} &
    \includegraphics[width=0.105\textwidth]{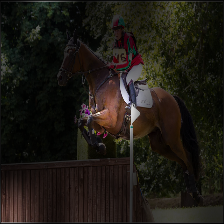} &
    \includegraphics[width=0.105\textwidth]{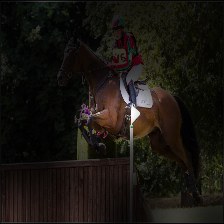} &
    \includegraphics[width=0.105\textwidth]{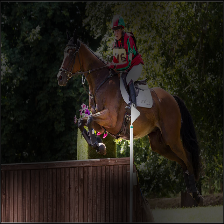} &
    \includegraphics[width=0.105\textwidth]{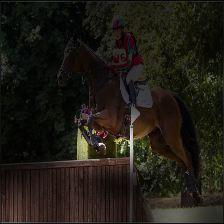} &
    \includegraphics[width=0.105\textwidth]{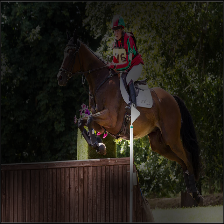} \\
    a & woman & is & riding & a & horse & over & an & obstacle \\

    \midrule

    \includegraphics[width=0.105\textwidth]{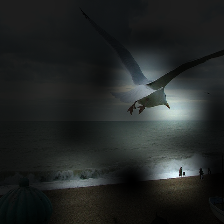} &
    \includegraphics[width=0.105\textwidth]{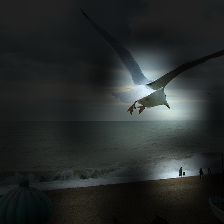} &
    \includegraphics[width=0.105\textwidth]{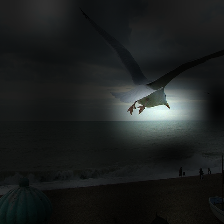} &
    \includegraphics[width=0.105\textwidth]{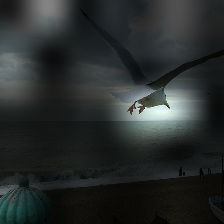} &
    \includegraphics[width=0.105\textwidth]{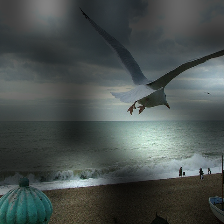} &
    \includegraphics[width=0.105\textwidth]{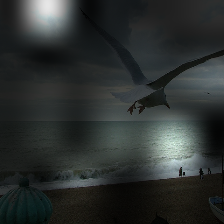} &
    \includegraphics[width=0.105\textwidth]{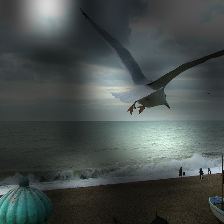} &
    \includegraphics[width=0.105\textwidth]{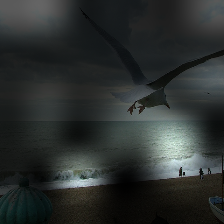} &
    \includegraphics[width=0.105\textwidth]{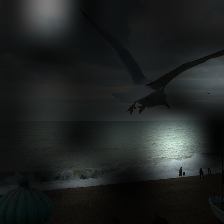} \\
    a & bird & flying & over & the & air & near & the & ocean \\

    \midrule

    \includegraphics[width=0.105\textwidth]{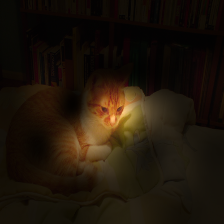} &
    \includegraphics[width=0.105\textwidth]{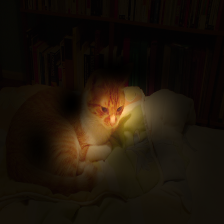} &
    \includegraphics[width=0.105\textwidth]{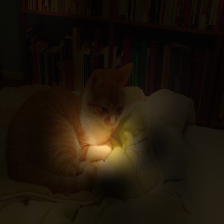} &
    \includegraphics[width=0.105\textwidth]{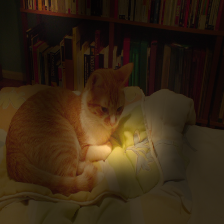} &
    \includegraphics[width=0.105\textwidth]{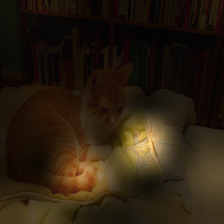} &
    \includegraphics[width=0.105\textwidth]{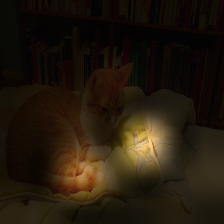} &
    \includegraphics[width=0.105\textwidth]{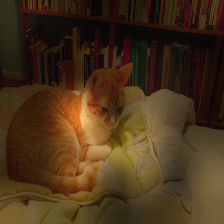} &
    \includegraphics[width=0.105\textwidth]{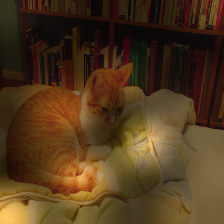} &
    \includegraphics[width=0.105\textwidth]{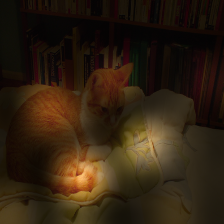} \\
    a & cat & laying & on & a & bed & in & a & bookshelf \\

    \midrule

    \includegraphics[width=0.105\textwidth]{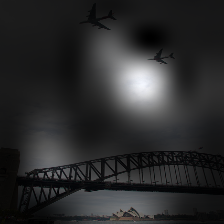} &
    \includegraphics[width=0.105\textwidth]{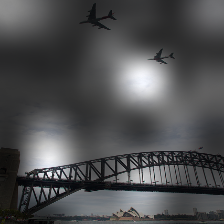} &
    \includegraphics[width=0.105\textwidth]{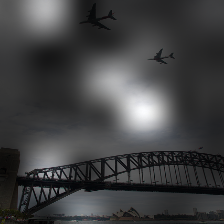} &
    \includegraphics[width=0.105\textwidth]{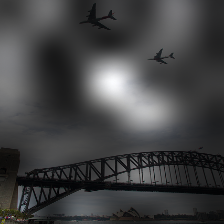} &
    \includegraphics[width=0.105\textwidth]{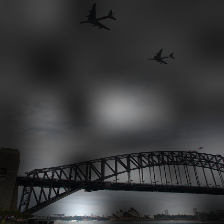} &
    \includegraphics[width=0.105\textwidth]{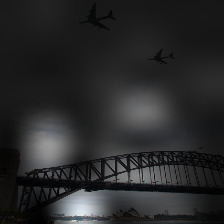} &
    \includegraphics[width=0.105\textwidth]{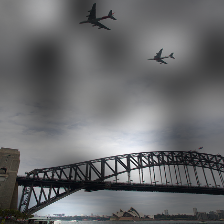} &
    \includegraphics[width=0.105\textwidth]{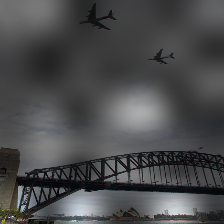} &
    \includegraphics[width=0.105\textwidth]{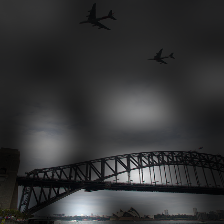} \\
    an & airplane & flying & over & a & body & of & a & river \\

    \midrule

    \includegraphics[width=0.105\textwidth]{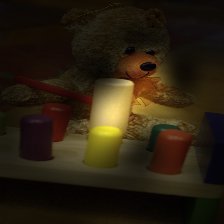} &
    \includegraphics[width=0.105\textwidth]{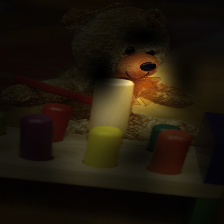} &
    \includegraphics[width=0.105\textwidth]{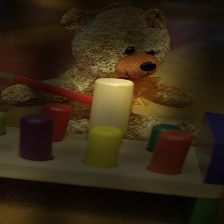} &
    \includegraphics[width=0.105\textwidth]{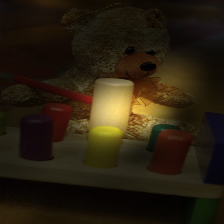} &
    \includegraphics[width=0.105\textwidth]{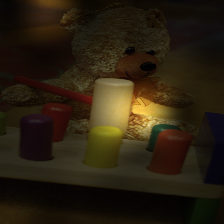} &
    \includegraphics[width=0.105\textwidth]{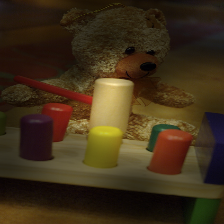} &
    \includegraphics[width=0.105\textwidth]{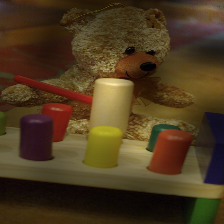} &
    \includegraphics[width=0.105\textwidth]{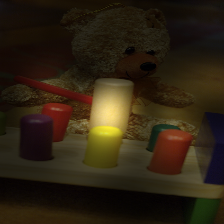} &
    \includegraphics[width=0.105\textwidth]{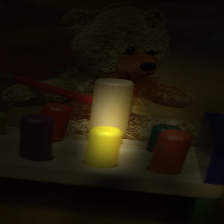} \\
    a & teddy & bear & sitting & in & front & of & orange & juice \\

    \midrule

    \includegraphics[width=0.105\textwidth]{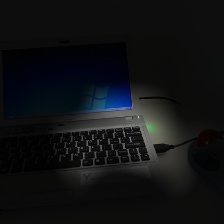} &
    \includegraphics[width=0.105\textwidth]{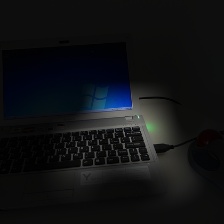} &
    \includegraphics[width=0.105\textwidth]{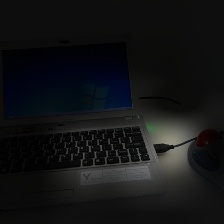} &
    \includegraphics[width=0.105\textwidth]{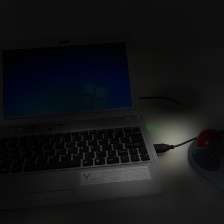} &
    \includegraphics[width=0.105\textwidth]{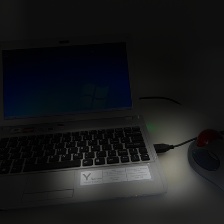} &
    \includegraphics[width=0.105\textwidth]{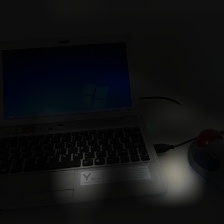} &
    \includegraphics[width=0.105\textwidth]{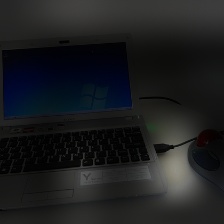} &
    \includegraphics[width=0.105\textwidth]{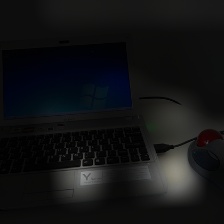} &
    \includegraphics[width=0.105\textwidth]{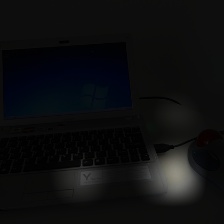} \\
    a & laptop & computer & sitting & on & top & of & a & desk \\

    \midrule

    \includegraphics[width=0.105\textwidth]{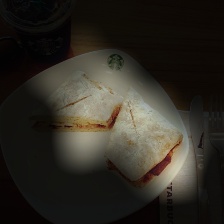} &
    \includegraphics[width=0.105\textwidth]{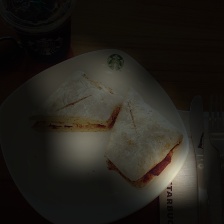} &
    \includegraphics[width=0.105\textwidth]{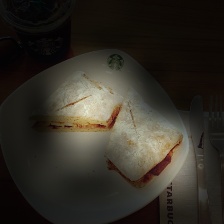} &
    \includegraphics[width=0.105\textwidth]{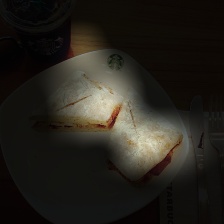} &
    \includegraphics[width=0.105\textwidth]{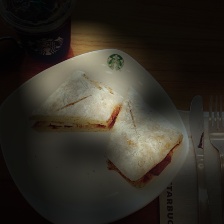} &
    \includegraphics[width=0.105\textwidth]{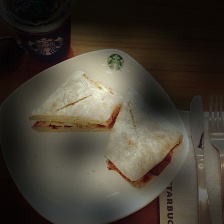} &
    \includegraphics[width=0.105\textwidth]{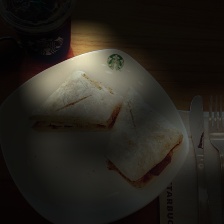} &
    \includegraphics[width=0.105\textwidth]{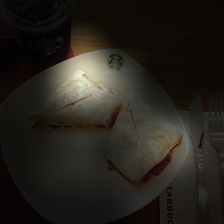} &
    \includegraphics[width=0.105\textwidth]{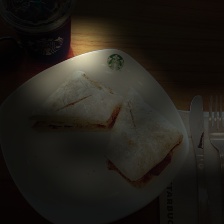} \\
    a & plate & with & a & sandwich & and & cup & of & coffee \\

  \end{tabularx}
  \caption{
    Attention visualizations per time step for predicted caption.
    We decode captions from the forward transformer of $L = 1, H = 512$ \virtex{} model using beam search.
  }
  \label{fig:supp_preds_all}
\end{figure*}

\begin{figure*}
    \newcolumntype{Y}{>{\centering\arraybackslash}X}
    \centering
    \footnotesize\ttfamily
    \setlength\tabcolsep{1pt}
    \begin{tabularx}{\textwidth}{YYYYY}

        \includegraphics[width=0.184\textwidth]{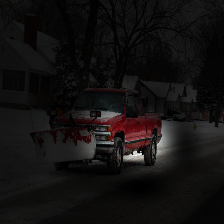} &
        \includegraphics[width=0.184\textwidth]{figures/attention_maps/16439_8.png} &
        \includegraphics[width=0.184\textwidth]{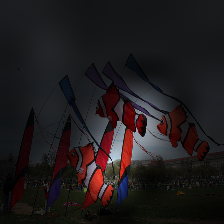} &
        \includegraphics[width=0.184\textwidth]{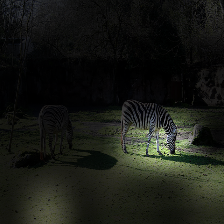} &
        \includegraphics[width=0.184\textwidth]{figures/attention_maps/18833_8.png} \\
        a red \attention{truck} driving down a snow covered road &
        a laptop computer sitting on top of a \attention{desk} &
        a group of \attention{kites} being flown in the park &
        two zebras are \attention{grazing} in a fenced in area &
        a cat laying on a pair of blue \attention{shoes} \\

        \midrule

        \includegraphics[width=0.184\textwidth]{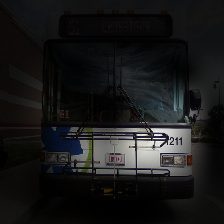} &
        \includegraphics[width=0.184\textwidth]{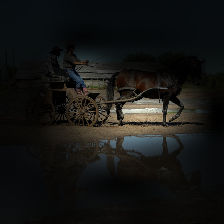} &
        \includegraphics[width=0.184\textwidth]{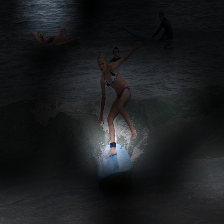} &
        \includegraphics[width=0.184\textwidth]{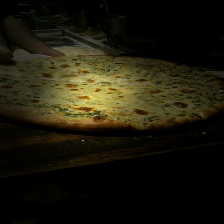} &
        \includegraphics[width=0.184\textwidth]{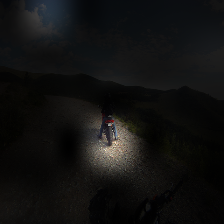} \\
        a \attention{bus} parked at the side of the road &
        a \attention{horse} drawn carriage being pulled by two horses &
        a woman on a wave \attention{board} in the ocean &
        a pizza on a cutting board on a \attention{pizza} &
        a person riding a \attention{motorcycle} on a dirt road \\

        \midrule
    
        \includegraphics[width=0.184\textwidth]{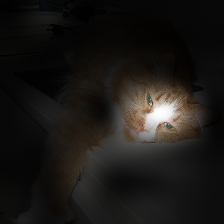} &
        \includegraphics[width=0.184\textwidth]{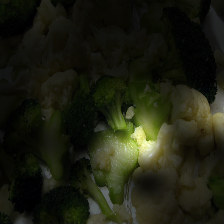} &
        \includegraphics[width=0.184\textwidth]{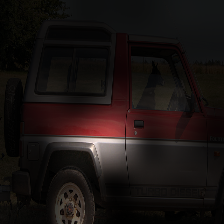} &
        \includegraphics[width=0.184\textwidth]{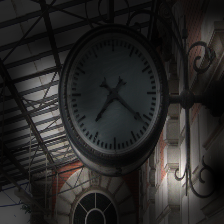} &
        \includegraphics[width=0.184\textwidth]{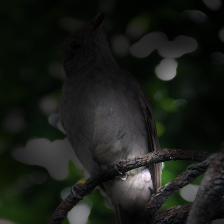} \\
        an orange and white \attention{cat} laying on a desk &
        a bowl of broccoli and \attention{cauliflower} in a lot &
        a dog in the back of a \attention{red} truck &
        a clock hanging from the ceiling in the \attention{ceiling} &
        a bird perched on top of a \attention{tree} branch \\

        \midrule

        \includegraphics[width=0.184\textwidth]{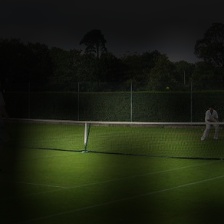} &
        \includegraphics[width=0.184\textwidth]{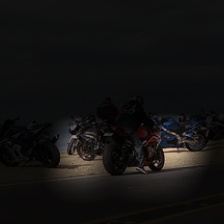} &
        \includegraphics[width=0.184\textwidth]{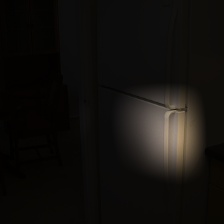} &
        \includegraphics[width=0.184\textwidth]{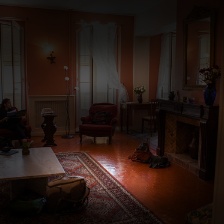} &
        \includegraphics[width=0.184\textwidth]{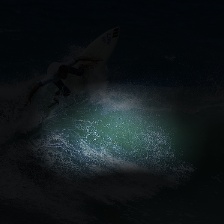} \\
        a group of people playing tennis on a tennis \attention{court} &
        a group of people riding \attention{motorcycles} down the road &
        a white \attention{refrigerator} freezer sitting in a kitchen next to a table &
        a living \attention{room} filled with furniture and a fireplace &
        a person on a surfboard riding a \attention{wave} in the ocean \\

        \midrule

        \includegraphics[width=0.184\textwidth]{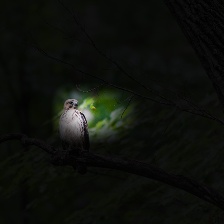} &
        \includegraphics[width=0.184\textwidth]{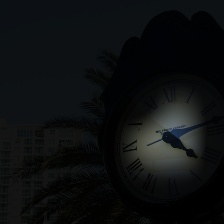} &
        \includegraphics[width=0.184\textwidth]{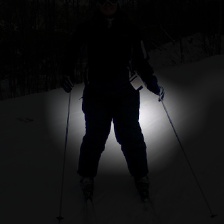} &
        \includegraphics[width=0.184\textwidth]{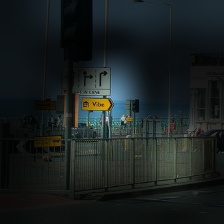} &
        \includegraphics[width=0.184\textwidth]{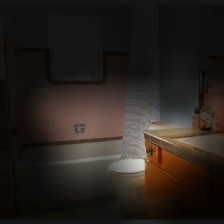} \\
        a \attention{bird} sitting on a branch of a tree &
        a \attention{clock} on a building with a clock on it &
        a woman on \attention{skis} in the side of a snow &
        a street \attention{sign} on it's edge of the road &
        a bathroom with a sink and \attention{toilet}, toilet \\

    \end{tabularx}
    \caption{
        We decode captions from the forward transformer of $L = 1, H = 512$ \virtex{} model using beam search.
        For the highlighted word, we visualize the decoder attention weights overlaid on the input image.
    }
    \label{fig:supp_preds_one}
\end{figure*}

In \Cref{fig:supp_preds_all} and \Cref{fig:supp_preds_one}, we show more qualitative examples showing decoder attention weights overlaid on input images, similar to \Cref{subsec:captioning}.
All captions are decoded from $L = 1, H = 512$ \virtex{} model using beam search.
We normalize the attention masks to $[0, 1]$ to improve their contrast for better visibility.

\end{appendices}

\end{document}